\documentclass[final,3p,times]{elsarticle}

\usepackage{amssymb}
\usepackage{amsmath}
\usepackage{algorithm}
\usepackage{algorithmic}
\usepackage{booktabs} % for professional tables

\renewcommand{\d}{\,\mathrm{d}}
\renewcommand{\vec}[1]{\boldsymbol{#1}}

\setkeys{Gin}{draft=false}

\journal{Applied Mathematical Modelling}

\begin{document}

\begin{frontmatter}



\title{Gradient-Free Score-Based Sampling Methods with Ensembles} %% Article title


\author[label1]{Bryan Riel\corref{contrib}}
\author[label2]{Tobias Bischoff\corref{contrib}} %% Author name

\affiliation[label1]{organization={Zhejiang Universitiy},
        addressline={briel@zju.edu.cn},
	city={Hangzhou},
	country={China}
}
\affiliation[label2]{organization={Aeolus Labs},%Department and Organization
            city={San Francisco, CA},
            country={United States}
}

\cortext[contrib]{Authors contributed equally}

%% Abstract
\begin{abstract}
Recent developments in generative modeling have utilized score-based methods coupled with stochastic differential equations to sample from complex probability distributions. However, these and other performant sampling methods generally require gradients of the target probability distribution, which can be unavailable or computationally prohibitive in many scientific and engineering applications. Here, we introduce ensembles within score-based sampling methods to develop gradient-free approximate sampling techniques that leverage the collective dynamics of particle ensembles to compute approximate reverse diffusion drifts. We introduce the underlying methodology, emphasizing its relationship with generative diffusion models and the previously introduced Föllmer sampler. We demonstrate the efficacy of the ensemble strategies through various examples, ranging from low- to medium-dimensionality sampling problems, including multi-modal and highly non-Gaussian probability distributions, and provide comparisons to traditional methods like the No-U-Turn Sampler. Additionally, we showcase these strategies in the context of a high-dimensional Bayesian inversion problem within the geophysical sciences. Our findings highlight the potential of ensemble strategies for modeling complex probability distributions in situations where gradients are unavailable. 
\end{abstract}


%\item Ensemble strategies enable gradient-free score-based sampling with reverse diffusion
%\item Adaptive importance sampling enhances ensemble score estimates
%\item Periodic resampling and antithetic sampling reduce estimation variance
%\item Prior-informed reverse diffusion improves high-dimensional sampling accuracy

%% Keywords

\end{frontmatter}



\section{Introduction}
\label{introduction}

Gradient-free methods for sampling from probability distributions are necessary in scenarios where gradient information is unavailable or computationally prohibitive. Such scenarios are common in Bayesian inverse problems involving complex, black-box simulators, legacy scientific code, or models with inherently non-differentiable components. Unfortunately, many performant samplers such as the No U-Turn Sampler (NUTS) or the Metropolis Adjusted Langevin Algorithm (MALA) require gradient information to fully realize their advantages \cite{roberts1996mala, hoffman2014nuts}. Recent investigations have utilized transport-based sampling methods to generate samples from probability distributions through stochastic differential equations (SDEs), e.g., \cite{huang2021schrodinger, vargas2023bayesian}. Diffusion models, developed within the field of generative modeling, can also be viewed as solving SDEs linking unknown data distributions to known prior distributions \cite{song2021scorebased}. In this context, \citet{huang2021schrodinger} suggested a Monte Carlo approximation of the Föllmer drift to sample from probability distributions \cite{follmer2005entropy, follmer2006time}, showing that this approach results in an approximation to the solution of the Schrödinger bridge problem.

We build on these ideas and introduce ensemble strategies that leverage the collective dynamics of a particle ensemble to approximate the score function (gradient of the log probability density function with respect to data) within a reverse diffusion process without actually needing to compute gradients. We develop these strategies with the aim to sample from arbitrary probability distributions known only up to a normalization constant, which is common in Bayesian inference applications. Compared to the aforementioned Föllmer sampler, our goal is to reduce the number of evaluations of the target probability distribution. In the setting of Bayesian inverse problems, this goal helps to reduce forward model evaluations, thereby offering an efficient sampling technique. More concretely, we introduce an importance sampling Monte Carlo estimator for the score function of a forward diffusion process in order to sample from a probability distribution using the associated reverse diffusion process. Moreover, we exploit the flexibility in the choice of the forward diffusion process to design samplers that are useful at medium- to high-dimensionality.

This work details the foundational methodology of our ensemble strategies, emphasizing its connection with generative diffusion models and the pre-existing Föllmer sampler. We assess the performance of our method through a range of diverse examples, encompassing low- to medium-dimensional Bayesian inference problems, including multi-modal and non-elliptical distributions. Our comparative analysis with Markov Chain Monte Carlo (MCMC) methods that utilize gradient information (e.g., \cite{roberts1996mala, hoffman2014nuts}) highlights the effectiveness of ensemble strategies in modeling complex probability distributions when gradients are not available. Furthermore, we demonstrate the practical application of our method in Bayesian inversion problems, particularly in the geophysical sciences where the dimensionality of the problem at hand can be of medium to large size, showcasing its potential for sophisticated probabilistic modeling.

\subsection{Related Work}
\label{related_work}
This paper builds upon existing research in stochastic processes, particularly focusing on the integration of Schrödinger-Föllmer approaches and particle and ensemble methods in computational physics (see, for example, \cite{reich2019data} or \cite{calvello2022ensemble} for recent overviews). Recent contributions adjacent to this area also include the work by \citet{boffi2023probability} on the probability flow solution of the Fokker-Planck equation and the study by \citet{bunne2022recovering} on recovering stochastic dynamics through Gaussian Schrödinger bridges. Further, \citet{dai2012global} explored aspects of global optimization using Schrödinger-Föllmer diffusion. \citet{huang2021schrodinger} and \citet{jiao2021convergence} provided an initial analysis of the error and convergence properties of Schrödinger-Föllmer samplers in non-convex settings and introduced ideas this work was inspired by. Building on this, \citet{zhang2022path} and \citet{vargas2023bayesian} demonstrated the use of neural networks to improve the efficiency of Föllmer drift estimates in high-dimensional settings, and \citet{debortoli2023diffusion} related the Schrödinger-Föllmer perspective to score-based generative models. In the area of ensemble-based gradient-free sampling techniques, Ensemble Kalman Inversion (EKI; \cite{schillings2017convergence}), Ensemble Kalman Sampling (EKS; \cite{garbuno2020interacting}), and related variants have been developed to address situation where gradients are unavailable. See \citet{chada2022review} or the aforementioned \citet{calvello2022ensemble} for overviews. Furthermore, ensemble based adaptive importance sampling techniques, such as the ensemble transport methods developed by \citet{cotter2019ensemble}, share conceptual similarities with our approach by utilizing the statistics of the current particle ensemble to dynamically construct more effective importance sampling distributions or transport steps.

\subsection{Our Contributions}
\label{our_contribution}
This work introduces ensemble strategies to calculate estimates of score functions. Our main contribution is an importance sampling Monte Carlo estimator that reuses the samples being generated in a reverse diffusion process to define an importance sampling distribution. This approach has the advantage of keeping the number of Monte Carlo samples necessary for estimating the score function of the reverse diffusion process low. Furthermore, we conduct extensive experiments with a hierarchy of problems, comparing our method against MCMC methods that utilize gradient information to improve sample efficiency (e.g., NUTS by \citet{hoffman2014nuts}). Lastly, we explore various parameters influencing performance, such as noise schedules, importance sampling strategies, and ensemble sizes, providing insights into its application in complex systems modeling. We also show how flexibility resulting from the choice of forward diffusion process can lead to performant algorithms that efficiently leverage prior information.

\section{Foundations of Score-based Sampling with Diffusion Models}
\label{theory}
This section introduces the theoretical foundations for what follows, establishing the mathematical framework and principles that guide our methodology. We first provide a brief overview of score-based sampling with diffusion models and then address some of the strategies that make our methods work for the demonstrated experiments.

% Need to mention needing ability to evaluate (un-normalized) likelihoods; also mention how SF processes have controlled dynamics that reach the target distribution in finite time

\subsection{Forward \& Reverse Diffusion Processes}
\label{heat_semi_group}
This section introduces the fundamental concepts of diffusion processes as they are used in this paper. Going forward, we consider forward diffusion processes of Ornstein-Uhlenbeck form
\begin{equation}
    \d \vec{x} = -\vec{b}_t\left(\vec{x} -\vec{\mu}\right) \d t + \vec{g}_t \d \vec{W}_t, 
    \label{eqn_fwd_process}
\end{equation}
where $\vec{b}_t$ is a time-dependent linear operator, $\vec{\mu}$ is the mean vector the process reverts to, and $\vec{g}_t$ is a time-dependent scaling matrix (often referred to as the drift and diffusion operators, respectively). The forward diffusion can be viewed as the incremental addition of noise to a data distribution. The associated reverse diffusion process \cite{anderson1982reverse} is given by
\begin{equation}
    \d \vec{x} = -\left(\vec{b}_t\left(\vec{x} -\vec{\mu}\right)+\vec{g}_t\vec{g}_t^T \vec{s}_t(\vec{x})\right) \d t + \vec{g}_t \d \vec{W}_t,
    \label{eqn_rev_process}
\end{equation}
where $\vec{s}_t$ is the time-dependent score of the solution of the forward diffusion equation and where Equation~\eqref{eqn_rev_process} is integrated backwards in time. Thus, once the score function is known for all $t$, samples can be generated from the original data distribution using the reverse diffusion SDE. Importantly, the score function allows for sampling from un-normalized probability distributions, which is imperative for applications like Bayesian inference \cite{hyvarinen2005score}. This forward-reverse duality has been leveraged in numerous studies within the areas of generative machine learning to generate samples from high dimensional data distributions, e.g., \cite{ song2021scorebased, song2019generative} and follow-on studies. As we demonstrate later with numerical examples, the diffusion framework provides flexibility in the construction of the forward and reverse processes (via selection of $\vec{b}_t$, $\vec{\mu}$, and $\vec{g}_t$), allowing for tuning of the sampling process for a given probability distribution.

For a given initial probability distribution $p_0$ that we want to sample from, we can write down an exact expression for the score related to the process in Equation~\eqref{eqn_fwd_process}:
\begin{subequations}
    \begin{align}
        \vec{s}_t(\vec{x}) &= \nabla_{\vec{x}} \log p_t(\vec{x}), \\
        p_t(\vec{x}) &=  \int \kappa_t(\vec{x}|\vec{x}') p_0(\vec{x}') \d \vec{x}', \\
        \kappa_t(\vec{x}|\vec{x}') &= \mathcal{N}\left(\vec{\mu}_t(\vec{x}'),\vec{\Sigma}_t\right),
    \end{align}
\end{subequations}
where $p_t$ is the solution to the Fokker-Planck equation associated with Equation~\eqref{eqn_fwd_process} and the kernel function $\kappa_t$ is a Gaussian corresponding to the associated transition probability (i.e., Green's function) of the Ornstein-Uhlenbeck process. The kernel mean and covariance can be derived analytically as \cite{gardiner1985handbook}
\begin{subequations}
    \begin{align}
        \vec{\mu}_t(\vec{x}') = \left(1 - \exp\left(-\int_0^{t'} \vec{b}_{t'} \d t'\right) \right)\vec{\mu} + \exp\left(-\int_0^t \vec{b}_{t'} \d t'\right) \vec{x}', \\
        \vec{\Sigma}_t = \int_0^t \exp\left(-\int_{t'}^t \vec{b}_{t''} \d t''\right) \vec{g}_{t'}\vec{g}_{t'}^T \exp\left(-\int_{t'}^t \vec{b}_{t''}^T \d t''\right) \d t'.
    \end{align}
\end{subequations}
Now consider a 1D Ornstein-Uhlenbeck process defined by $\mu = 0$, $g_t = \sigma$, and $b_t = b$ is constant in time. In this case, $\mu_t(x') = x' \exp(-t)$ and $\Sigma_t = (\sigma^2 / (2b)) (1 - \exp(-2bt))$ such that as $t \rightarrow \infty$, $\mu_t \rightarrow 0$ and $\Sigma_t \rightarrow \sigma^2 / (2b)$. In other words, the solution to the forward diffusion equation interpolates between the initial distribution $p_0$ and a normal distribution with zero mean and a fixed variance. For standard Brownian motion, $\mu_t(x') = x'$ and $\Sigma_t = \sigma^2 t$, i.e. the forward diffusion approaches a normal distribution with infinite variance as $t \rightarrow \infty$.

\subsection{Score-based Sampling in Bayesian Inference}
In a Bayesian context, when $p_0$ is a posterior distribution, and if the prior distribution is a normal distribution with a given mean and covariance, then this approach can be used to approximately transport samples from the prior to the posterior over a finite time span $0 \leq t \leq 1$ by choosing $\vec{b}_t$, $\vec{\mu}$, and $\vec{g}_t$ appropriately. Given the kernel of the forward diffusion process, one can then use the reverse diffusion equation to draw samples from $p_0$. This behavior can be used to tune the selection of the forward process parameters for a given sampling problem. However, the main challenge lies in the fact that the score $\vec{s}_t$ is only available as a potentially high-dimensional integral. In the next sections, we introduce an approach that uses an ensemble of samples to define an importance sampling estimator for the score, thereby alleviating this problem for low- to medium-dimensional problems.

\section{Ensemble Strategies for Gradient-Free Score Estimation and Sampling}

Building upon the diffusion framework, this section introduces ensemble-based strategies designed to estimate the score function and enable efficient sampling from target distributions without requiring gradient information. We detail the core Monte Carlo score estimation and various techniques to enhance its accuracy and applicability, particularly for complex and high-dimensional problems.

\subsection{Estimating the Score Function}
\label{foellmer_drift}
\begin{figure}[ht]
\begin{center}
\includegraphics[width=0.85\textwidth]{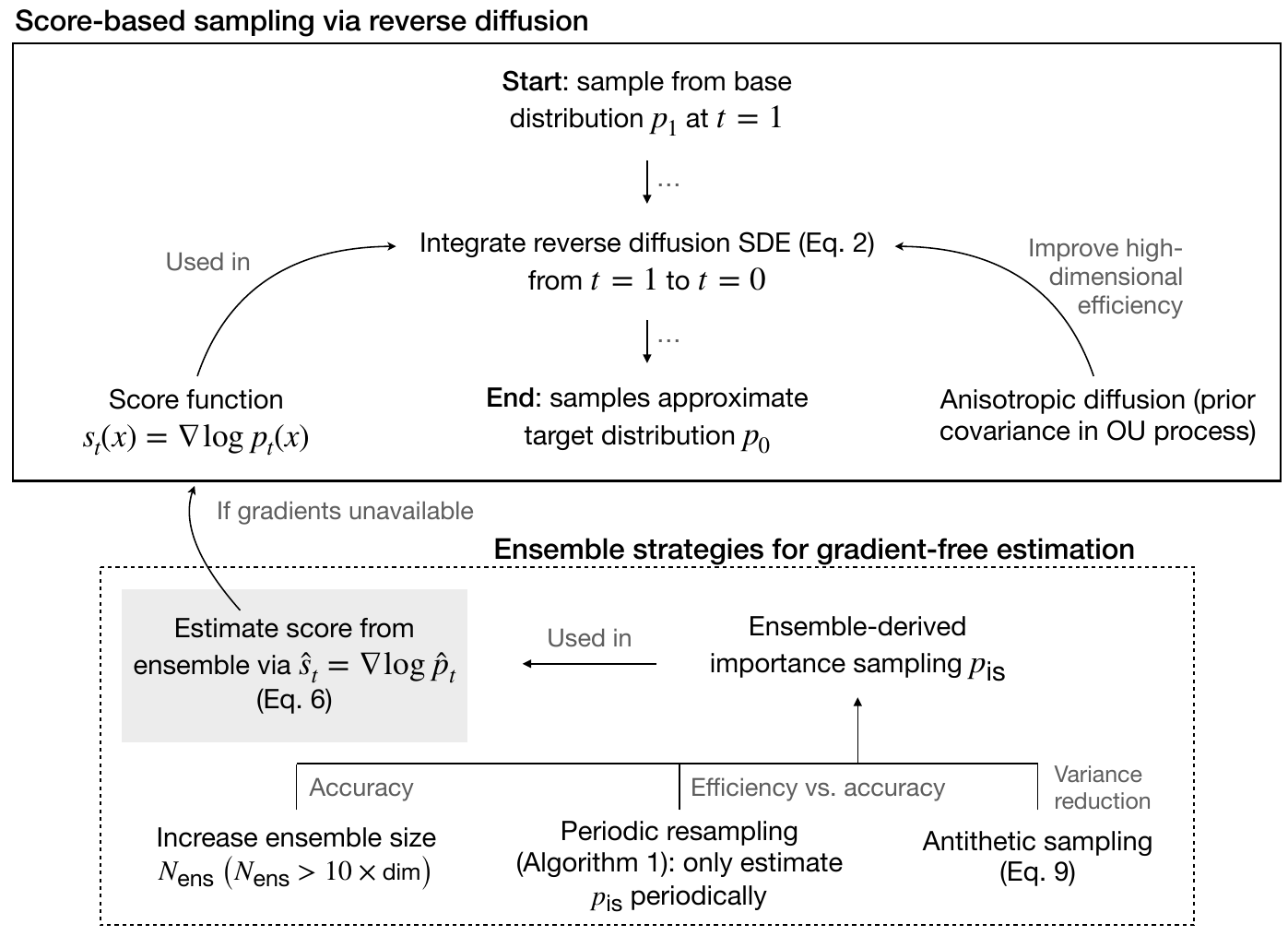}
\caption{Flowchart illustrating the core reverse diffusion sampling process and the ensemble-based estimation framework used to replace the score function when gradients cannot be computed, highlighting the roles of the different variance reduction and efficiency strategies.}
\label{fig:flowchart}
\end{center}
\end{figure}

A key aspect of our ensemble strategies is the estimation of the score within the reverse diffusion process, which is essential for accurately simulating particle trajectories in Equation~\eqref{eqn_rev_process}. This section outlines our approach to estimating the score function. To that end, we introduce an importance sampling distribution $p_{\rm{is}}$ (e.g., \cite{elvira2022advances}) such that we can rewrite the solution to the forward diffusion equation as
\begin{equation}
    p_t(\vec{x}) = \int \kappa_t(\vec{x}|\vec{x}') \frac{p_0(\vec{x}')}{p_{\rm{is}}(\vec{x}')} p_{\rm{is}}(\vec{x}') \d \vec{x}'.
\end{equation}
We now proceed to define a Monte Carlo approximation of the form
\begin{equation}
    \hat{p}_t(\vec{x}) = \frac{1}{N} \sum_{i=1}^N \kappa_t(\vec{x}|\vec{x}'_i) \frac{p_0(\vec{x}'_i)}{p_{\rm{is}}(\vec{x}'_i)}, \hspace{.25cm} \vec{x}'_i \sim p_{\rm{is}}, 
    \label{eqn_follmer_drift_mc}
\end{equation}
where $N$ is the number of Monte Carlo samples drawn. Here, our goal is to define $p_{\rm{is}}$ in such a way that the reverse diffusion process generates good samples from the posterior distribution $p_0$ even though we use an approximation for the score. When actually computing the estimate $\hat{\vec{s}}_t(\vec{x}) = \nabla_{\vec{x}} \log \hat{p}_t(\vec{x})$ for the score function, we follow the approach of \citet{vargas2023bayesian} and use the logsumexp trick to dramatically improve the numerical stability of the score estimation. A flowchart summarizing the ensemble score-based sampling procedure (and variance reduction strategies for the Monte Carlo estimator $\hat{p}_t$ discussed in the next section) is shown in Figure \ref{fig:flowchart}.

\subsection{Variance Reduction Strategies}
\label{variance_reduction}
This section focuses on the implementation of various variance reduction strategies that are essential for enhancing the accuracy and efficiency of the sample generation process. These strategies can be grouped into: i)  general variance reduction strategies that we apply to all sampling problems (Sections \ref{sec_imp_sampling}-\ref{section_periodic_resampling}); and ii) problem-specific strategies that utilize additional information to improve sample quality (Section \ref{subsection_ou_fwd}).

\subsubsection{Importance Sampling via Sample Ensemble}
\label{sec_imp_sampling}
One key aspect of our approach is to simultaneously consider ensembles of samples that are being generated. That is to say, we consider not just a single sample being generated in the reverse diffusion process, but rather an ensemble $\{\vec{x}_i \in \mathbb{R}^D\}_{1 \hdots N_{\rm{ens}}}$, where $N_{\rm{ens}}$ is the ensemble size. At $t=1$, the initial conditions for the reverse diffusion process are drawn from a normal distribution. Then, as $t$ is reduced from $1$ to $0$, the normal distribution samples are transformed into samples from the distribution $p_0$. As the ensemble progresses, we can define an ensemble mean and covariance as
\begin{subequations}
    \begin{align}
        \vec{\mu}_{\rm{ens}} &= \frac{\sum_{i=1}^{N_{\rm{ens}}} \vec{x}_i}{N_{\rm{ens}}}, \\
        \vec{\Sigma}_{\rm{ens}} &= \frac{\sum_{i=1}^{N_{\rm{ens}}} \left(\vec{x}_i - \vec{\mu}_{\rm{ens}}\right)\left(\vec{x}_i - \vec{\mu}_{\rm{ens}}\right)^T}{N_{\rm{ens}}}, 
    \end{align}
    \label{eqn_esfs_is}
\end{subequations}
and an associated importance sampling distribution
\begin{equation}
    p_{\rm{is}} = \mathcal{N}(\vec{\mu}_{\rm{ens}}, \vec{\Sigma}_{\rm{ens}}).
\end{equation}
Utilizing this ensemble-based $p_{\rm{is}}$ therefore relies on the adequacy of the normal distribution as an importance sampling distribution. In practice, we also set the number of Monte Carlos samples $N$ equal to the ensemble size $N_{\rm{ens}}$. We outline an alternative importance sampling distribution using a multiple importance sampling strategy based on the ensemble of samples in \ref{appendix_mis}. We find that both distributions work similarly well for the examples presented in this paper. The actual variance reduction achieved is directly related to the ensemble size; that is, as the ensemble size tends to infinity, the score estimator becomes exact (see \ref{appendix_convergence_analysis} for a convergence analysis). In practice, we find that we do not need to actually draw samples from $p_{\rm{is}}$ when calculating estimators for $\hat{p}_t$ or $\hat{\vec{s}}_t$. Instead, we find that evaluating the Monte Carlo approximation at the ensemble members gives equally reasonable results.

\subsubsection{Antithetic sampling}
We also find in our experiments that an antithetic sampling strategy can be beneficial to enhance the quality of the importance sampling estimator. The antithetic version of the importance sampling estimator can be written as
\begin{align}
    \hat{p}_t(\vec{x}) &= \frac{\sum_{i=1}^{N_{\rm{ens}}} \kappa_t(\vec{x}|\vec{x}'_i) \frac{p_0(\vec{x}'_i)}{p_{\rm{is}}(\vec{x}'_i)}}{2N_{\rm{ens}}} + \frac{\sum_{i=1}^{N_{\rm{ens}}} \kappa_t(\vec{x}|-\vec{x}'_i) \frac{p_0(-\vec{x}'_i)}{p_{\rm{is}}(-\vec{x}'_i)}}{2N_{\rm{ens}}}.
    \label{eqn:antithetic}
\end{align}
This variance reduction strategy is common in Bayesian contexts (e.g., \cite{geweke1988antithetic}). The impact on the quality of the score estimator is shown in Figure~\ref{fig_ensemble_size}.

\subsubsection{Periodic Resampling Strategy}
\label{section_periodic_resampling}
Since the variance of the estimator $\hat{p}_t$ depends on $p_{\rm{is}}$, which itself depends on the statistics of the ensemble $\{\vec{x}_i\}$, we can refine $\hat{p}_t$ during the reverse diffusion by updating $p_{\rm{is}}$ using the statistics of $\{\vec{x}_i\}$ at discrete resampling times, $t_r$, where $0 < t_{\rm{r}} < 1$. After each resampling time, $\hat{p}_t$ is held fixed until the next resampling time. By increasing the number of resampling times, $N_{\rm{r}}$, during reverse diffusion, $p_{\rm{is}}$ is adapted to remain in the proximity of $p_0$ during the diffusion process, which prevents the likelihood ratio $p_0(\vec{x}') / p_{\rm{is}}(\vec{x}')$ from getting too small to effectively perform importance sampling \cite{agapiou2017importance}. Importantly, $p_0$ is only evaluated when updating $\hat{p}_t$, so there is a trade-off between increasing $N_{\rm{r}}$ to improve the score estimate while decreasing $N_r$ to limit the total number of evaluations of $p_0$. For the experiments shown here, $N_r$ is generally set to between 10 and 30. Overall, this strategy is analogous in spirit to MCMC methods that utilize proposal distributions that adapt to the statistics of the Markov chains during sampling (e.g., \cite{haario2001mcmc,minson2013bayesian}). The outline of this process is shown in Algorithm~\ref{alg:esfs}.
\begin{algorithm}[ht]
   \caption{Resampling Strategy}
   \label{alg:esfs}
    \begin{algorithmic}
       \STATE {\bfseries Input:} initial ensemble $\vec{x}_i \in \mathbb{R}^D$, $i \in \{1, \hdots, N_{\rm{ens}}\}$
       \STATE mean vector $\mathbb{\vec{\mu}}$, scaling matrix $\vec{g}_t$
       \STATE number of resampling steps $N_{\rm{r}} \Rightarrow \Delta t_{\rm{r}} = \frac{1}{N_{\rm{r}}}$ 
       \STATE time step size $\Delta t$
       \FOR{$r=1$ {\bfseries to} $N_{\rm{r}}$}
       \STATE update $p_{\rm{is}}$ using $\{\vec{x}_i\}_{i=1\hdots N_{\rm{ens}}}$
       \STATE update $\hat{p}_t$ and hence $\hat{\vec{s}}_t$ using antithetic samples
       \STATE $t_{\rm{start}} \leftarrow 1 - (r-1) \Delta t_{\rm{r}}$
       \STATE $t_{\rm{end}} \leftarrow 1 - r \Delta t_{\rm{r}}$
       \FOR{$i=1$ {\bfseries to} $N_{\rm{ens}}$}
            \STATE solve reverse diffusion equation:
            \STATE $\vec{x}_i \leftarrow \mathrm{odesolve}(\vec{x}_i; t_{\rm{start}}, t_{\rm{end}}, \mathbb{\vec{\mu}}, \vec{g}_t, \hat{\vec{s}}_t, \Delta t)$
       \ENDFOR
       \ENDFOR
       \STATE {\bfseries Return:} $\{\vec{x}_i\}_{i=1\hdots N_{\rm{ens}}}$
    \end{algorithmic}
\end{algorithm}
By recomputing $p_{\rm{is}}$ only periodically, we reduce the number of $p_0$ evaluations, which is beneficial when $p_0$ is expensive to evaluate. This situation is common in complex black box simulations settings where not only gradients are unavailable, but evaluations of the black box are also computationally expensive. With this approach, the total number of forward evaluations in Algorithm~\ref{alg:esfs} is limited to $N_{\rm{ens}}N_{\rm{r}}$. For example, $N_{\rm{r}}=10$ resampling steps and an ensemble size $N_{\rm{ens}} = 100$ would result in $1000$ evaluations of $p_0$.

\subsubsection{Anisotropic Forward Diffusion}
\label{subsection_ou_fwd}

For certain classes of problems where some information about the covariance structure of the parameters is available a priori, we can employ a localization-type strategy. Specifically, the flexible form of Equation~\eqref{eqn_fwd_process} allows us to choose the scaling matrix of the forward diffusion to be a localization matrix defined as the Cholesky-decomposed prior covariance. In this way, we can ``push'' the diffusion process to stay ``close'' to the dominant subspace defined by the prior covariance. The implicit assumption is that the posterior will follow a similar covariance structure as defined by the prior. In essence, this localization strategy is a kind of linear dimensionality reduction strategy where we assume that we know the relevant subspace a priori (e.g., \cite{webber2023localized}). As a result, this strategy becomes more useful as the dimensionality of the problem increases. In the examples presented here, we utilize an Ornstein-Uhlenbeck forward process for the forward diffusion process with drift and diffusion operators defined as
\begin{subequations}
  \label{eqns_ou_ops}
  \begin{align}
    \vec{b}_t &= \theta (\vec{\mu} - \vec{x}_t), \\
    \vec{g}_t &= \text{Cholesky}(\alpha \vec{\Sigma}_{\text{prior}}),
  \end{align}
\end{subequations}
which evolves samples from $p_0$ to a multivariate Gaussian distribution with mean $\vec{\mu}$ and covariance matrix $\alpha \vec{\Sigma_{\text{prior}}}$ over a time scale $\theta > 0$. Here, $\alpha$ is a scalar inflation factor for scaling the prior covariance in order to optionally inflate the initial ensemble. We choose the magnitude of $\theta$ such that as $t \rightarrow 1$ the process has converged to $\mathcal{N}(\vec{\mu}, \alpha \vec{\Sigma}_{\text{prior}})$ in a distributional sense. The reverse diffusion process can then be determined by simply plugging $\vec{b}_t$ and $\vec{g}_t$ into Equation~\eqref{eqn_rev_process}.

\begin{figure}[ht]
\begin{center}
\includegraphics[width=1.0\textwidth]{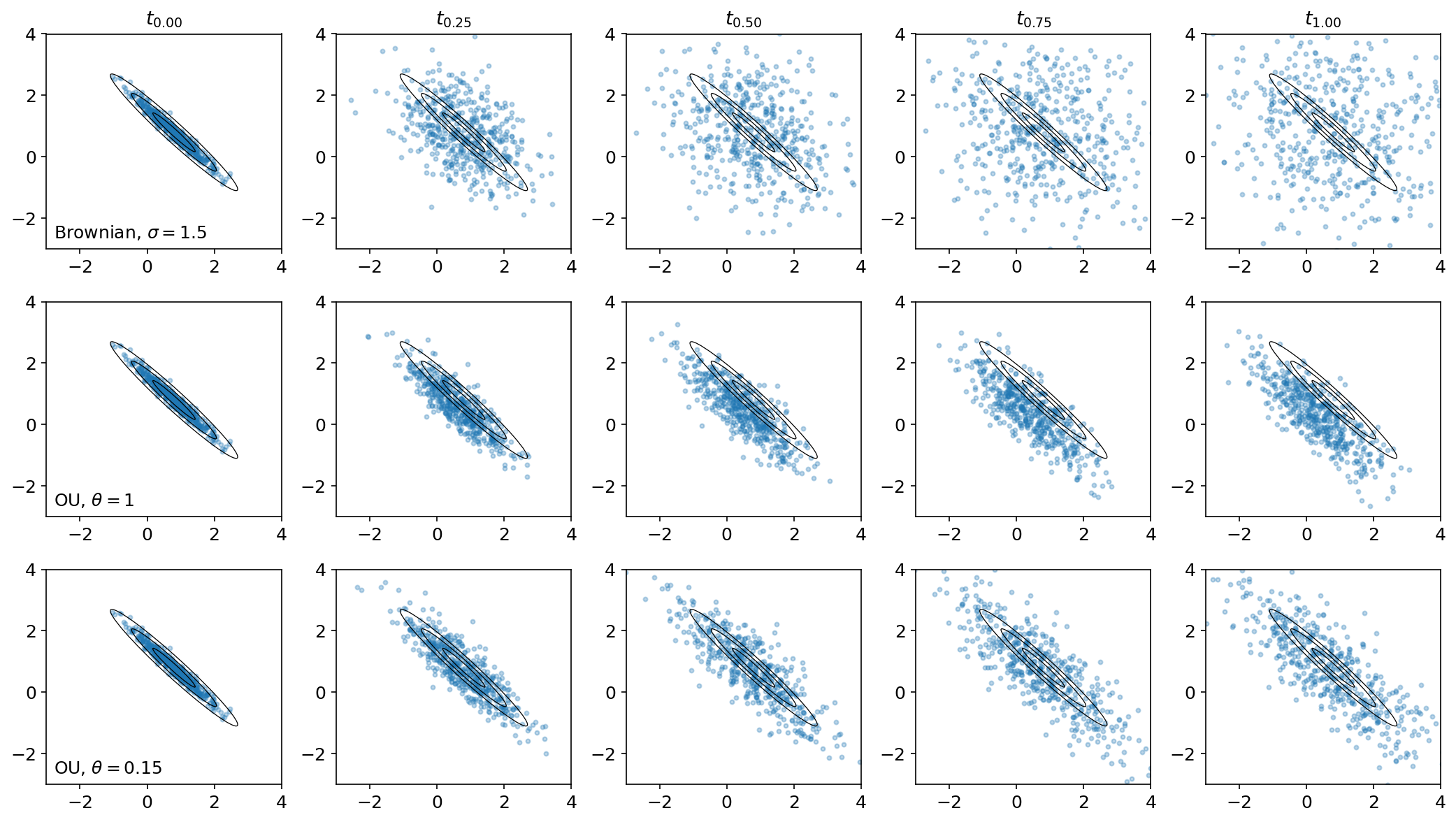}
\caption{Simulated forward diffusion for different functional forms of the forward process. The initial sample ($t_{0.00}$, left) is generated from a multivariate Gaussian with $\mu_0 = [0.8, 0.8]$ and $\vec{\Sigma}_0 = [[0.4, -0.39], [-0.39, 0.4]]$. \textbf{Top:} Brownian process with zero drift and $\sigma = 1.5$. \textbf{Middle:} Ornstein-Uhlenbeck process with $\theta = 1$, $\mu = 0$, and $\vec{\Sigma} = [[0.5, -0.4], [-0.4, 0.5]]$. \textbf{Bottom:} Same as the middle Ornstein-Uhlenbeck process but with $\theta = 0.15$.}
\label{fig_ou_fwd_2d}
\end{center}
\end{figure}

The utility of the Ornstein-Uhlenbeck forward process can be illustrated with a simple 2D example of forward diffusion from a multivariate normal distribution. For scalar Brownian motion, the samples will take random walks in each dimension with equal variance. For higher-dimensional problems, this behavior can cause many samples to move towards low probability regions of the target distribution. With the Ornstein-Uhlenbeck process, the random walks are constrained to the covariance structure imposed by $\vec{g}_t$, which for certain problems can keep samples in the relevant regions of the probability space to improve the estimate of the Föllmer drift (Figure \ref{fig_ou_fwd_2d}). In practice, we also tune $\theta$ to control the relative importance of $\mu$ during the reverse diffusion. For $\theta = 1$, the samples will stay close to $\mu$, which can be un-desirable when the target mean is expected to differ non-negligibly from $\mu$. For the higher-dimensional experiments in this work, we find $\theta = 0.1 - 0.2$ to provide the best accuracy.

\subsection{Comparison with Ensemble Kalman Methods}
Sampling and optimization methods based on Ensemble Kalman methods use an ensemble covariance matrix to estimate the gradient of a log probability density function that serves as the drift within a Langevin-type SDE, see for example \cite{garbuno2020interacting} for an easy-to-digest exposition. Usually, one can show that these methods become exact samplers in a Gaussian setting and convergence results can be obtained, for example, in convex and mean field settings \cite{calvello2022ensemble, schillings2017convergence, garbuno2020interacting}. Score-based reverse diffusion samplers, as presented here, have the added advantage that they become exact samplers for nearly any target distribution as the ensemble size and resampling times tend to infinity (see Section~\ref{section_lorenz63} for a numerical comparison of the two methods for the Lorenz63 system). However, score-based methods, as presented here, have the added disadvantage that they fight an uphill battle against the curse of dimensionality, discussed further in Sections~\ref{moderate_dim_problems} and~\ref{high_dim_problems}. This situation is not surprising as estimating the score function in high dimensions using Monte Carlo sampling alone is a losing proposition because controlling the variance of the estimator becomes increasingly difficult. In short, we expect similar sampling accuracy between our method and Ensemble Kalman methods for near-Gaussian probability distributions and improved accuracy for our method for non-Gaussian distributions.

\section{Experiments}
\label{applications}

In this section, we demonstrate the effectiveness of our proposed ensemble strategies for score-based sampling by conducting a series of comprehensive tests across a variety of sampling problems, ranging from low- to high-dimensional scenarios. We provide a comparison against performant gradient-based MCMC techniques and demonstrate improvements in sampling accuracy without the need for gradient information.

\subsection{Two-dimensional Problems}
\label{low_dim_problems}
\begin{figure*}[ht]
\begin{center}
\includegraphics[width=1.0\textwidth]{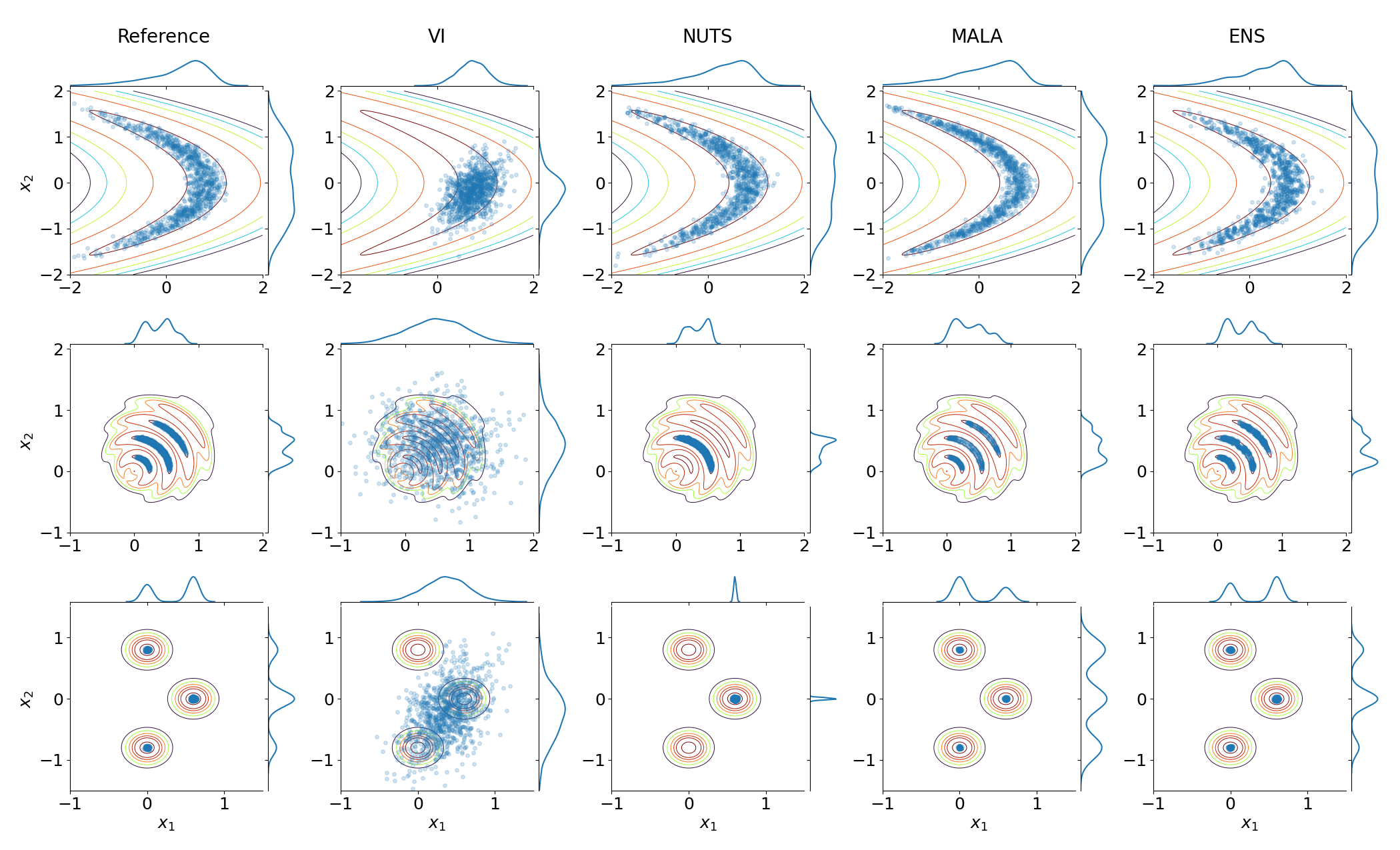}
\caption{Comparison of different methods for drawing random samples from three different two-dimensional probability distributions. For each distribution, the joint plot shows contours of log probability along with the random samples while the marginal plots show kernel density estimates of the marginal distributions. The leftmost column shows direct samples from the distributions, while the rest of the columns represent samples from a variational inference approximation (VI), the No U-Turn Sampler (NUTS), a Metropolis-adjusted Langevin Algorithm (MALA), and our approach (ENS).}
\label{fig_2d_comparison}
\end{center}
\end{figure*}

We first apply our ensemble strategy to a set of two-dimensional Bayesian toy problems that exhibit posterior distributions with varying degrees of multimodality and non-ellipticity, which generally provide challenges to many MCMC techniques for sampling probability distributions. As shown in Figure~\ref{fig_2d_comparison}, these problems include the well-known ``banana''-shaped distribution, a distribution with neighboring high-probability ridges (adapted from \cite{morzfeld2018iterative}), and a mixture of three Gaussian distributions with well-separated means. The parameters used for each experiment are specified in Table~\ref{table_esfs_2d}.
\begin{table}[t]
\caption{Parameters for the examples shown in Figure \ref{fig_2d_comparison}. For these experiments, the forward process has zero drift (i.e., $\vec{b}_t = \vec{\mu} = 0$), and the diffusion $\vec{g}_t = (\sigma_{\rm{min}}^{1/p} + t (\sigma_{\rm{max}}^{1/p} - \sigma_{\rm{min}}^{1/p}))^p$, where $p = 5$ \cite{karras2022elucidating}. MIS here stands for the multiple importance sampling approach from Appendix~\ref{appendix_mis}. Additionally, $N_{\rm{ens}} = 1000$, $N_r = 10$, and the initial time step size is specified by $\Delta t_{\rm{init}}$.}
\label{table_esfs_2d}
\vskip 0.15in
\begin{center}
\begin{small}
\begin{sc}
\begin{tabular}{lcccr}
\toprule
Distribution & $\sigma_{\rm{min}}$, $\sigma_{\rm{max}}$ & $p_{\rm{is}}$  & $\Delta t_{\rm{init}}$ \\
\midrule
Banana & (0.01, 1) & Mixture (MIS) & 0.005 \\
Ridged & (0.005, 1) & Mixture (MIS) & 0.005 \\
Mixture & (0.005, 1) & Gaussian & 0.0025 \\
\bottomrule
\end{tabular}
\end{sc}
\end{small}
\end{center}
\vskip -0.1in
\end{table}
For comparison, we generate samples from these target distributions using three different techniques: 1) approximation of the distribution with a 2D multivariate Gaussian with parameters estimated via minimization of the reverse KL-divergence (i.e., variational inference (VI)); 2) NUTS; and 3) MALA. Note that for NUTS, a single chain is used for sampling, whereas for MALA, multiple chains are initialized with random states and only the final states are used as samples of the target distribution. In \ref{appendix_gradient_free_algo}, we also show the performance of two established gradient-free sampling algorithms: i) Population Monte Carlo (PMC), and ii) Ensemble Slice Sampling (ESS).
\begin{table}[ht]
\centering
\begin{tabular}{lccc}
\hline
Method & Banana & Ridged & Mixture \\
\hline
VI & 39.280 $\pm$ 9.100 & 6.698 $\pm$ 2.033 & 14.822 $\pm$ 5.038 \\
NUTS & 3.981 $\pm$ 3.360 & 20.435 $\pm$ 3.032 & 16.620 $\pm$ 3.682 \\
MALA & 7.143 $\pm$ 4.894 & 0.891 $\pm$ 0.672 & 6.768 $\pm$ 3.397 \\
ESS & 4.436 $\pm$ 3.127 & 0.592 $\pm$ 0.437 & 16.623 $\pm$ 3.667 \\
PMC & 2.969 $\pm$ 2.521 & 0.504 $\pm$ 0.444 & 9.352 $\pm$ 3.794 \\
\textbf{ENS} & \textbf{3.102 $\pm$ 2.616} & \textbf{0.531 $\pm$ 0.389} & \textbf{1.945 $\pm$ 1.835} \\
\hline
\end{tabular}
\caption{Comparison of \textbf{energy distance} (mean and standard deviation) for the two-dimensional experiments using different sampling methods.}
\label{tab:energy_distance_transposed}
\end{table}

As expected, the non-ellipticity of all three distributions present difficulties for methods that rely on or are related to Gaussian approximations (which includes the aforementioned Ensemble Kalman methods). Moreover, the reverse KL-divergence cost function used for VI encourages mode-seeking approximations, which can miss substantial areas of high probability for skewed or multi-modal distributions \cite{blei2017variational}. Alternatively, both NUTS and MALA perform well for the ``banana'' distribution, but for the second target distribution, NUTS with a single chain has a tendency to draw samples from a single mode or ridge of high probability, even when the ridges are not well-separated. Multiple runs of NUTS with different random initializations would likely improve sampler exploration in this case. MALA is most similar to our method in that a diffusion process controls the random walk of samples that are initially randomly distributed, which permits samples to reach multiple regions of high probability. However, Langevin dynamics can still suffer from slow mixing when modes are separated by low density regions (as is the case for the third distribution), causing MALA to generate samples with incorrect relative densities for the mixture of Gaussians distribution \cite{song2019generative}.

To quantify sampling accuracy, an energy-based distance measurement (\ref{appendix_energy}) that estimates the statistical distance between two sets of random vectors is computed for the samples relative to samples from the true distribution (Table \ref{tab:energy_distance_transposed}). Our method is able to correctly sample all three distributions, demonstrating its ability to handle both non-ellipticity and multi-modality. These results broadly confirm the results obtained in \cite{huang2021schrodinger} where a Föllmer sampler was used to generated results from two-dimensional distributions, albeit their approach differs from ours in several ways (e.g., different Monte Carlo estimators, number of Monte Carlo samples, etc.)

\subsubsection{Influence of Ensemble Size}
\label{low_dim_ensemble_size}
As shown in Equations~\eqref{eqn_follmer_drift_mc} and~\eqref{eqn_esfs_is}, the ensemble size affects the accuracy of the Monte Carlo estimate for the score function. Incorrect drift estimates will cause the sample trajectories to move toward incorrect locations in parameter space. In practice, the minimum ensemble size for sufficiently accurate drift estimates will depend on a multitude of factors, including the dimensionality of the probability distribution, the choice of importance sampling distribution, and the design of the forward process (\ref{appendix_convergence_analysis}). We follow the practice of Ensemble Kalman methods by starting with an ensemble size that is 10 times the dimensionality of the probability distribution \citep[e.g.,][]{schillings2017convergence,garbuno2020interacting}. We then compare the final posterior samples with those obtained from an ensemble size that is 1.5--2 times larger. If the samples exhibit substantially different marginal distributions, then we increase the ensemble size further.

\begin{figure}[ht]
\begin{center}
\includegraphics[width=0.55\columnwidth]{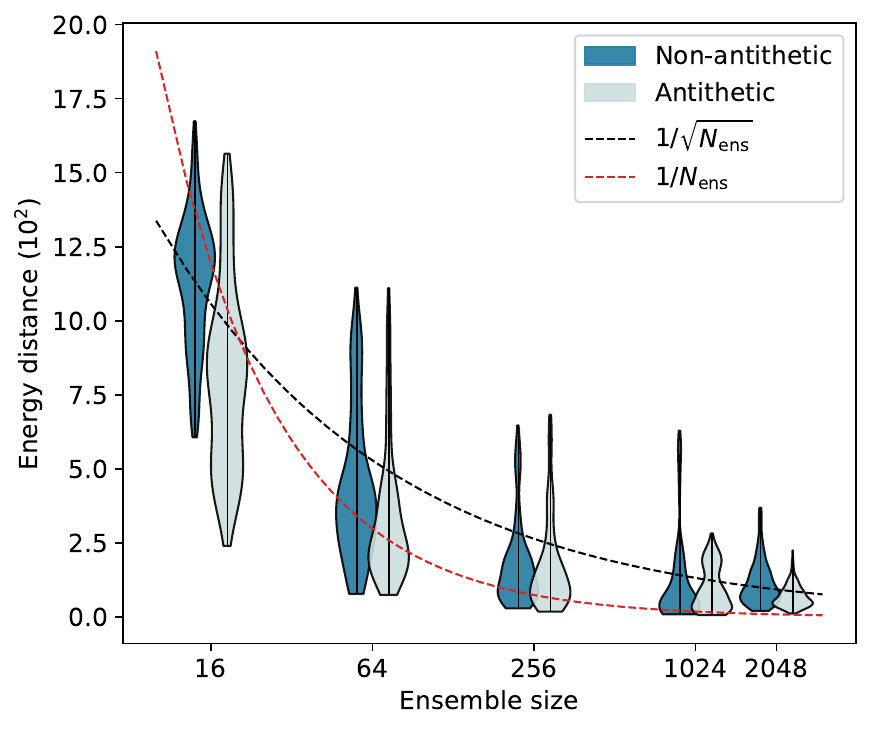}
\caption{Violin plots showing energy distance between samples generated by our methods and ground truth samples from a 5-$d$ Gaussian as a function of ensemble size, $N_{\rm{ens}}$, and dependent on whether an antithetic sampling strategy is used. At each ensemble size, energy distance was computed for $200$ random permutations of the samples. The violin plots are bounded by the 10th and 90th percentiles of the resulting energy statistics. Black and red dashed lines are the theoretical $1/\sqrt{N_{\rm{ens}}}$ and $1/N_{\rm{ens}}$ curves, respectively.}
\label{fig_ensemble_size}
\end{center}
\end{figure}

To explore this effect, we use our method to sample from a five-dimensional multivariate Gaussian distribution for different ensemble sizes and measure the energy distance relative to the reference 5-$d$ Gaussian. We find that the energy distance of our method is reduced by a factor of three when increasing $N_{\rm{ens}}$ from 16 to 64, with a further factor of two reductions with $N_{\rm{ens}}$ from 64 to 256 (Figure \ref{fig_ensemble_size}). These reductions scale roughly with $\sqrt{N_{\rm{ens}}}$, consistent with the convergence analysis in \ref{appendix_convergence_analysis}. For smaller ensemble sizes, the antithetic sampling strategy can significantly decrease sampling errors, while for larger ensembles, antithetic sampling can minimize the occurrence of high energy distances associated with inaccurate diffusion trajectories. These trends are broadly consistent across different target probability distributions, but the exact ensemble size at which accuracy will degrade will require experimentation on a case-by-case basis. More robust and automated selection of ensemble sizes remains an active area of research.

\subsubsection{Influence of Ensemble Initialization}
\label{section_ensemble_init}
\begin{figure}[ht]
\begin{center}
\includegraphics[width=1.0\textwidth]{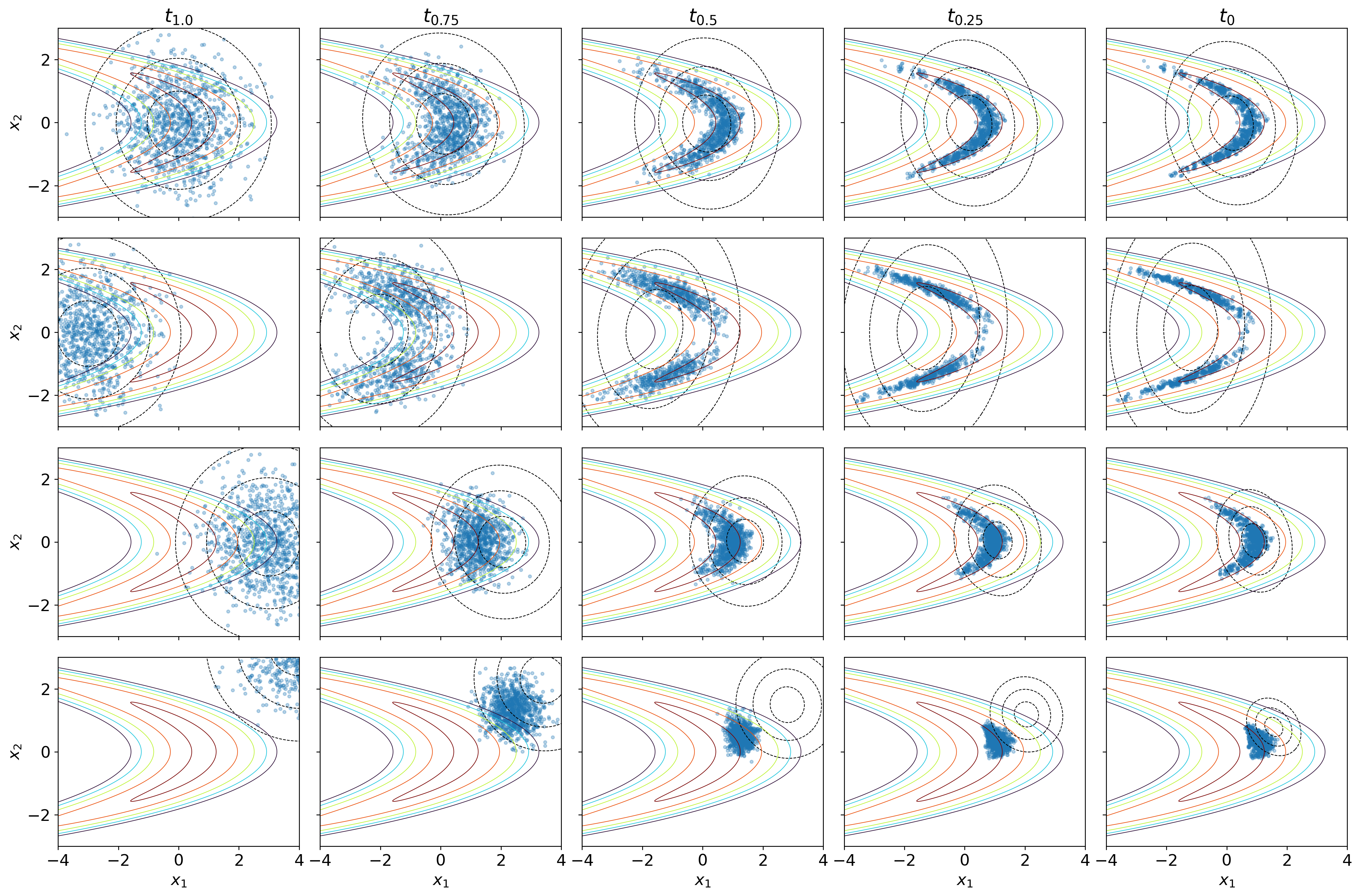}
\caption{Reverse diffusion trajectories for different initial ensembles for the banana probability distribution. The diffusion times span from $t = 1$ (left) to $t = 0$ (right), and the different rows correspond to different initial ensembles with varying offsets relative to the true distribution. Black dashed elliptical contours correspond to the importance sampling distribution, $p_{\text{is}}$, at the given diffusion times. In these examples, a Gaussian $p_{\text{is}}$ is used. Overall, ensembles with larger offsets will lead to undersampling of the distribution.}
\label{fig_rev_banana}
\end{center}
\end{figure}

Similar to the ensemble size, the seeding of the initial ensemble members will affect the accuracy of the estimated score function. Generally, the Monte Carlo estimate requires that a few ensemble members have sufficiently high initial likelihood values in order to estimate an accurate score using Equation~\eqref{eqn_follmer_drift_mc}. We can see this effect in Figure \ref{fig_rev_banana} where we sample from the 2D banana distribution with three different initial ensembles.
If only a few ensemble members have locations that lie close to the target probability distribution, the rest of the ensemble will evolve towards those locations, leading to undersampling of the distribution. However, even when the initial ensemble lies entirely outside of the distribution (e.g., the last row in Figure \ref{fig_rev_banana}), the variance of the importance sampling distribution can still introduce a sufficient dynamic range in likelihood values to allow the ensemble to evolve in the correct direction. Therefore, an iterative approach may be applied where the reverse diffusion ensemble is re-initialized with a bias determined from the mean of the ensembles from previous runs. In this way, the initial ensemble has greater overlap with the target distribution.

\subsection{Three-dimensional Lorenz63 Problem}
\label{section_lorenz63}

The Lorenz63 system is a well-known simplified model for atmospheric convection consisting of the following three ordinary differential equations:
\begin{align}
\frac{dy_1}{dt} &= \sigma (y_2 - y_1), \\
\frac{dy_2}{dt} &= y_1 (r - y_3) - y_2, \\
\frac{dy_3}{dt} &= y_1 y_2 - \beta y_3,
\end{align}
where $\sigma$, $r$, and $\beta$ are scalar system parameters that are normally assumed to be positive. The trajectories of $y_1$, $y_2$, and $y_3$ can be computed with a standard ODE solver with initial conditions sampled from a unit normal distribution. For certain combinations of $\sigma$, $r$, and $\beta$, the trajectories will exhibit chaotic behavior that is highly sensitive to the initial conditions. This behavior prevents the reliable computation of gradients of the system outputs with respect to system parameters, which ultimately prevents the use of gradient-based MCMC methods to estimate the parameter values from observations. Here, we use the values $\sigma = 10$, $r = 24.5$, and $\beta = 8/3$, which places the system near the transition between stable fixed points and chaotic behavior.

Following the approach of \citet{huang2022iterated}, we set up a Bayesian inference problem that aims to sample the posterior distribution of $\mathbf{x} = [\sigma, r, \beta]$ from observations consisting of time-averages of the first and second moments of the system outputs, $\mathbf{d} = [\overline{y_1}, \overline{y_2}, \overline{y_3}, \overline{y_1^2}, \overline{y_2^2}, \overline{y_3^2}]$. The time averages are computed as $\overline{y_i} = \frac{1}{t_f - t_i} \int_{t_i}^{t_f} y_i dt$, which incorporates a spin-up period of $t = t_i$ to eliminate transient effects from the initial conditions. The observations $\mathbf{d}$ are computed using $t_i = 30$ and $t_f = 200$, and a corresponding covariance matrix is estimated from samples of $\mathbf{d}$ computed with a rolling window of $t = 20$. During inference for $\mathbf{x}$, the forward model predictions for $\mathbf{d}$ are computed with $t_i = 30$ and $t_f = 50$.

\subsubsection{Sampling the Posterior Distribution}

The Bayesian inference problem formulates the un-normalized posterior distribution for $\mathbf{x}$ as:
\begin{align}
p(\vec{x} | \vec{d}) \sim p(\vec{d} | \vec{x}) p(\vec{x}),
\end{align}
where the first distribution on the right-hand side is the likelihood distribution and the second distribution is the prior distribution. Here, we use a multivariate normal distribution for the likelihood with the mean vector and covariance matrix as described previously. For the prior, we use an independent normal distribution, $p(\vec{x}) = \mathcal{N}(\vec{\mu}, \vec{\sigma^2 I})$ with $\vec{\mu} = [12, 25, 8/3]$ and $\vec{\sigma} = [2, 1, 1]$. In practical applications, since $\mathbf{x}$ is strictly positive, we should sample for $\log(\vec{x})$, which would lead to a near-Gaussian posterior $p(\log(\vec{x}) | \vec{d})$. Since one of our goals for this experiment is to highlight the difference in sampling behavior between our proposed method and Ensemble Kalman Sampling (EKS) \cite{garbuno2020interacting}, we opt to sample for $\mathbf{x}$, which will lead to non-Gaussian behavior for $p(\vec{x} | \vec{d})$. Furthermore, we multiply the covariance matrix by a factor of four when parameterizing the likelihood distribution in order to enhance the effect of the non-linear forward model on $p(\vec{x} | \vec{d})$.

For sampling $p(\vec{x} | \vec{d})$ with the reverse diffusion process, we prescribe a zero-drift forward diffusion process with $\vec{g}_t$ as in Table \ref{table_esfs_2d}, $\sigma_\text{min} = 0.05$, $\sigma_\text{max} = 1$, $\Delta t_\text{init} = 0.0025$, an MIS $p_\text{is}$, 20 resampling steps, and an ensemble size of $N_\text{ens} = 1000$. For EKS, in order to prescribe the same number of forward model evaluations as our method, we use an ensemble size of 1000 for 20 iterations. Additionally, we sample $p(\vec{x} | \vec{d})$ with Population Monte Carlo (PMC), which is well suited for low-dimensional, non-Gaussian distributions.

\begin{figure}[ht]
\begin{center}
\includegraphics[width=1.0\textwidth]{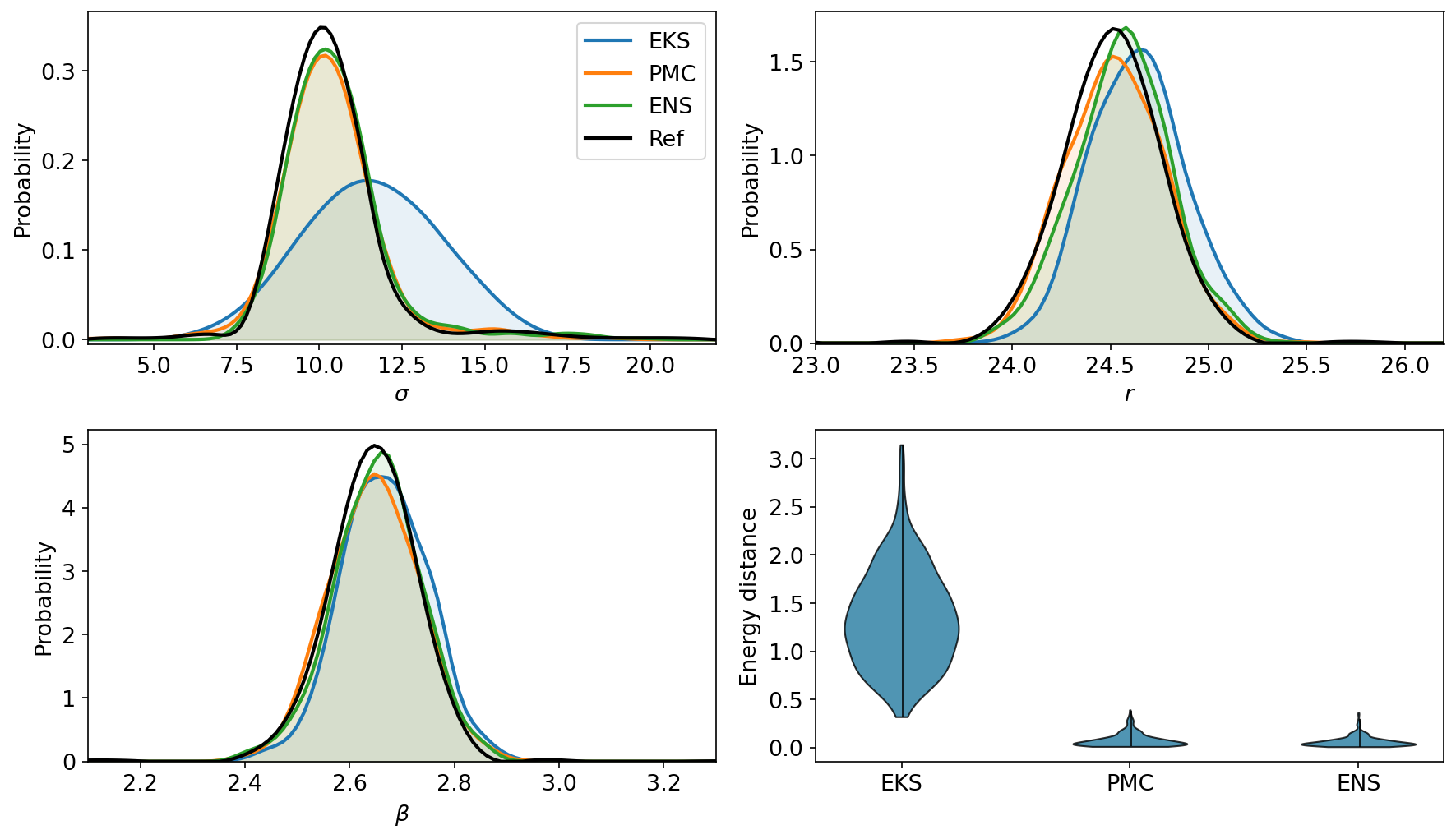}
\caption{\textbf{Marginal posterior kernel density estimates} for the three parameters ($\sigma$, $r$, $\beta$) of the Lorenz63 system. 1D marginals are computed from samples obtained with the Ensemble Kalman sampler (EKS; blue), Population Monte Carlo (PMC; orange), and the ensemble methods presented here (ENS; green). The true posterior probabilities are shown in black. Energy distances relative to the true posterior are shown in the bottom right plot. Overall, all methods perform well for sampling parameters that exhibit near-Gaussian marginal posterior distributions, while non-Gaussian behavior is more accurately sampled by the more flexible PMC and ensemble methods presented here.}
\label{fig_lorenz}
\end{center}
\end{figure}

The posterior samples derived from all three methods successfully encompass the true values (Figure \ref{fig_lorenz}). For the parameters $r$ and $\beta$, the marginal posterior is close to Gaussian, so the samples derived from all methods are nearly identical. However, the marginal posterior for $\sigma$ has a longer, low-probability tail for higher values. Since EKS utilizes Gaussian approximations for the posterior, the associated marginal samples for $\sigma$ have an inflated variance in order to span the tail, leading to oversampling of that region. In contrast, our score-based sampling method and PMC generate samples closer to the true posterior distribution. These results indicate that our proposed method is effective for Bayesian inference problems for highly non-linear systems that are infeasible to solve with gradient-based techniques.

\subsection{Moderate-dimensional Problems}
\label{moderate_dim_problems}

\begin{figure}[ht]
\begin{center}
\includegraphics[width=0.7\textwidth]{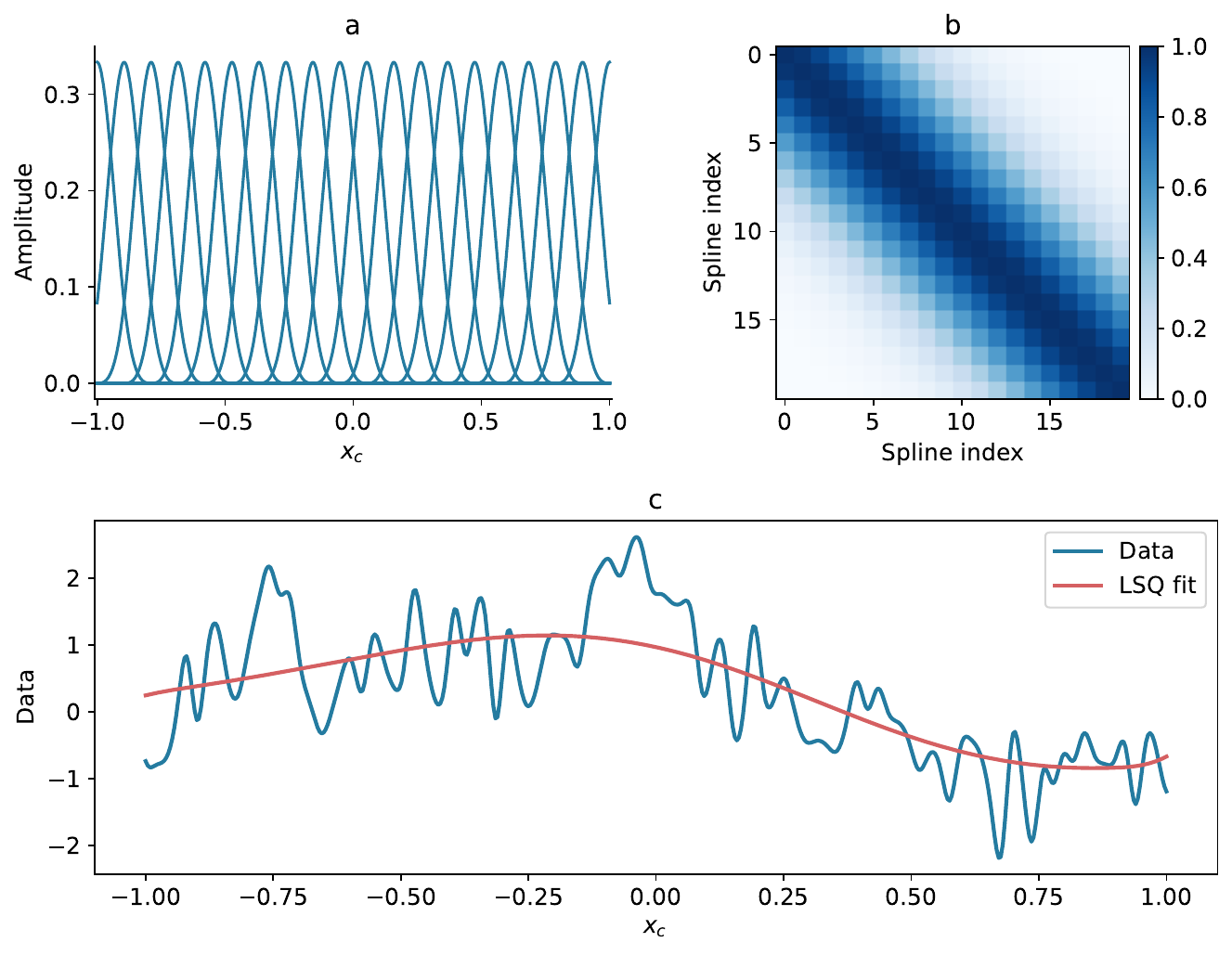}
\caption{\textbf{Top left (a):} B-splines used in the Bayesian linear regression example from Section~\ref{moderate_dim_problems}. \textbf{Top right (b)}: Prior covariance of spline coefficients in Bayesian linear regression example. \textbf{Bottom (c):} Synthetic noisy data and corresponding least squares fit used for reference solution.}
\label{fig_splines}
\end{center}
\end{figure}

As we expect the intrinsic variance of the Monte Carlo estimation of the score function to increase rapidly with increasing dimensionality, it is useful to investigate a moderate-dimensional problem ($D < 10$) to characterize the sampling behavior of our proposed methods and the utility of our variance reduction strategies. To that end, we formulate a 20-dimensional Bayesian linear regression problem corresponding to estimation of the coefficients, $\vec{x} \in \mathbb{R}^{20}$, of uniformly spaced B-splines placed in the columns of a linear operator $\vec{G} \in \mathbb{R}^{500\times20}$, given a data vector $\vec{d} \in \mathbb{R}^{500}$. The B-splines serve as smooth interpolating basis functions for reconstructing signals from noisy data (Figure \ref{fig_splines}a.) \cite{hetland2012multiscale}.
In this class of Bayesian regression problems, it is common to enforce a prior constraint that the spline coefficients, represented by the vector $\vec{x}$, are correlated over a given length scale, which can be represented by a squared-exponential Gaussian Process kernel $k(x_i, x_j) = \exp\left[-d(x_i, x_j)^2 /  (2L^2)\right]$, where $d(x_i, x_j)$ is the Euclidean distance and $L$ is the prescribed length scale (e.g., \cite{webber2023localized, jolivet20142013}). In this experiment, the B-splines are defined within a spatial domain of $[-1, 1]$, and the length scale $L$ is set to $0.5$.

With this setup, the aim is obtain samples from the posterior distribution of the spline coefficients given a set of (potentially noisy) observations. For a given data vector $\vec{d} = \vec{Gx} + \vec{\epsilon}$, where $\vec{\epsilon} \sim \mathcal{N}(\vec{0}, \sigma_d^2 \vec{I})$ is normally distributed noise with variance $\sigma_d$, and a multivariate Gaussian prior, $p(\vec{x}) \sim \mathcal{N}(\vec{0}, \vec{\Sigma}_{\rm{prior}})$, we can derive an analytic solution for $p(\vec{x} | \vec{d})$ as the least-squares estimate \cite{tarantola2005inverse}:
\begin{align}
\hat{\vec{x}} = \left(\vec{G}^T\vec{G} + \sigma_d^2 \vec{\Sigma}_{\text{prior}}^{-1}\right)^{-1}\vec{G}^T\vec{d}.
\end{align}
The corresponding posterior covariance matrix for $\hat{\vec{x}}$ is
\begin{align}
\vec{\Sigma}_{\hat{\vec{x}}} = \sigma_d^2 \left(\vec{G}^T\vec{G} + \sigma_d^2 \vec{\Sigma}_{\text{prior}}^{-1}\right)^{-1}.
\end{align}
Together, $\hat{\vec{x}}$ and $\vec{\Sigma}_{\hat{\vec{x}}}$ are used to construct a reference multivariate Gaussian distribution which we can use to assess the accuracy of various sampling methods. For this experiment, we generate $\vec{d}$ by sampling a reference $\vec{x}$ from a unit Gaussian distribution and multiplying by $\vec{G}$. We additionally add noise with a length scale of 0.05 and a variance of $\sigma_d = 2.0$ to obtain the final $\vec{d}$ used in the experiments.

\subsubsection{Sampling the Posterior Distribution}

For low-dimensional problems, the linear operator $\vec{b}_t$ and the scaling matrix $\vec{g}_t$ in Equation~\eqref{eqn_fwd_process} can be replaced with scalar terms such that the forward diffusion process evolves with equal variance in each dimension. As mentioned in Section \ref{subsection_ou_fwd}, as the dimensionality increases, isotropic random walks can cause ensemble members to move to areas of low probability, similarly to the standard Metropolis-Hastings algorithm. One commonly used strategy in score-based sampling is to convert the SDE in Equation~\eqref{eqn_rev_process} to a deterministic ODE (probability flow) \cite{song2021scorebased}. In this way, the only source of randomness is in the initial ensemble, which is then evolved using only the drift term in Equation~\eqref{eqn_rev_process}. However, since deterministic sampling continuously reduces noise levels during reverse diffusion \cite{karras2022elucidating}, the final samples may undersample $p_0$ and collapse to local modes. Moreover, another challenge with larger dimensionality is that the fraction of initial ensemble members $\{\vec{x}_i\}$ within low-probability regions of $p_0$ quickly approaches 1, which results in inaccurate Monte Carlo estimates of the score function and resulting drift.

\begin{figure*}[ht]
\begin{center}
\includegraphics[width=0.8\textwidth]{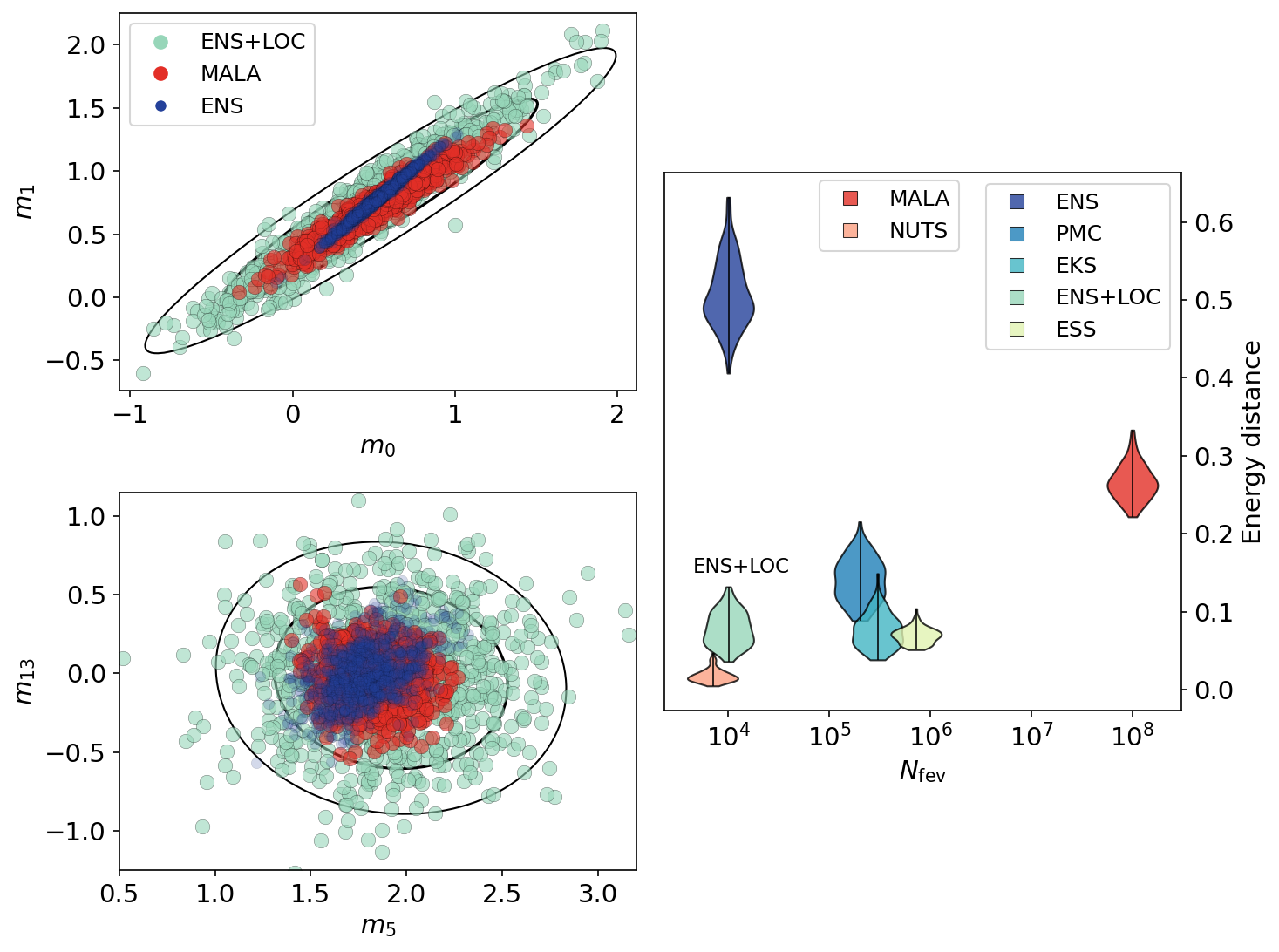}
\caption{Comparison of sampling methods for 20-D Bayesian linear regression. Left: \textbf{Marginal posterior samples} versus $3\sigma$ analytic uncertainty. Right: \textbf{Energy distance} vs. number of forward evaluations (blue-green=gradient-free, orange=gradient-based). Basic Ensemble method (ENS) underestimates variance, while other gradient-free methods (Particle Monte Carlo - PMC, Ensemble Slice Sampling - ESS) are computationally costly. ENS+LOC, using a prior-informed Ornstein-Uhlenbeck process, achieves accuracy comparable to Ensemble Kalman Sampling (EKS) and gradient-based methods while requiring fewer forward evaluations without gradients.}
\label{fig_20d}
\end{center}
\end{figure*}

As discussed in Section \ref{subsection_ou_fwd}, we can utilize an Ornstein-Uhlenbeck forward process to constrain the trajectories of the samples during reverse diffusion. Here, we construct the scale matrix $\vec{g}_t$ directly from the prior covariance (see Table~\ref{table_esfs_high_dim} for full specification of the Ornstein-Uhlenbeck parameters). This variance reduction strategy allows us to obtain samples from the target posterior distribution that have nearly the same accuracy as methods that utilize gradients of the log posterior (as measured by the energy distance between the generated samples and the analytic posterior distribution; see Figure~\ref{fig_20d}). While our method with a simple isotropic Gaussian diffusion term can successfully locate the mode of the posterior distribution, the sample variance is too low, leading to a high energy distance. MALA is able to increase the variance of the samples, albeit lower than that expected from the true posterior distribution. We note that MALA with fine-tuned Langevin dynamics could provide marginal improvement but would still be limited by its use of isotropic Brownian motion, which places it under the same limitations as our method with an isotropic diffusion term. Our method with an Ornstein-Uhlenbeck forward process performs nearly as well as NUTS, which requires gradients of the posterior distribution to fully explore the probability space. Moreover, compared to other gradient-free algorithms, our method achieves higher accuracy with significantly fewer forward evaluations, $N_{\text{fev}}$.
\begin{table}[t]
\caption{Parameters for the 20-$d$ and 100-$d$ problems utilizing the Ornstein-Uhlenbeck forward process with localization based on the prior covariance.}
\label{table_esfs_high_dim}
\vskip 0.15in
\begin{center}
\begin{small}
\begin{sc}
\begin{tabular}{lcccccr}
\toprule
Problem & $\theta$ & $\mu$ & $\alpha$ & $\Delta t_{\rm{init}}$ & $N_{\rm{ens}}$ & $N_r$ \\
\midrule
20-$d$ & 0.1 & 0 & $4^2$ & 0.002 & 1000 & 10 \\
100-$d$ & 0.05 & 0 & $5^2$ & 0.002 & 2000 & 30 \\
\bottomrule
\end{tabular}
\end{sc}
\end{small}
\end{center}
\vskip -0.1in
\end{table}

\subsection{High-dimensional Problems}
\label{high_dim_problems}

\begin{figure}[ht]
\begin{center}
\includegraphics[width=1.0\textwidth]{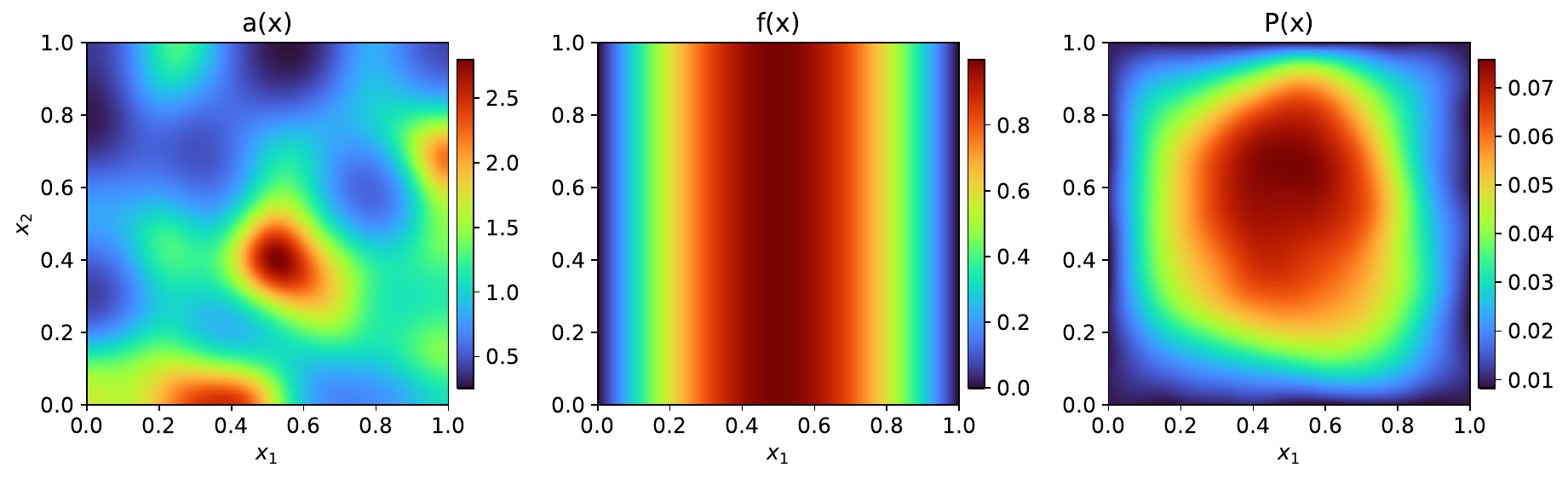}
\caption{Spatial fields with normalized units used in the pressure field PDE. The permeability field (a) and source field (b) lead to the pressure field (c) following Equation \ref{darcy_pde}. Noise with a standard deviation of 0.001 is added to the pressure field.}
\label{fig_pressure_fields}
\end{center}
\end{figure}

For this example, we aim to estimate a permeability field, $a(\vec{x})$, that leads to an observed pressure field, $P(\vec{x})$, in a porous medium resulting from a scalar field of fluid sources or sinks, $f(\vec{x})$. This system is governed by the following PDE for the pressure field $P(\vec{x})$ following the derivation of Darcy flow (e.g., \cite{garbuno2020interacting}):
\begin{subequations}
\begin{align}
\label{darcy_pde}
-\nabla \cdot \left(a(\vec{x}) \nabla P(\vec{x})\right) &= f(\vec{x}), \quad \vec{x} \in \mathcal{D}, \\ 
P(\vec{x}) &= 0, \qquad \vec{x} \in \partial \mathcal{D},
\end{align}
\end{subequations}
where $\mathcal{D}$ represents the interior of the modeling domain and $\partial \mathcal{D}$ represents the boundary. We prescribe $\mathcal{D}$ to span the domain $[0, 1]^2$ and assume that all fields have normalized units. We solve the above equations for a unit square mesh with Dirichlet boundary conditions using \texttt{firedrake} \cite{FiredrakeUserManual}. The ground truth $a(\vec{x})$ is modeled as  $a(\vec{x}) = a_0 e^{\theta(\vec{x})}$, where $a_0 = 0.6$ and $\theta(\vec{x})$ is a random Gaussian field with a length scale of 0.15. The simulated fields are shown in Figure \ref{fig_pressure_fields}.

The inverse problem aims to recover $\theta(\vec{x})$, which ensures that $a(\vec{x})$ is strictly positive. We simplify the problem by assuming that both $f(\vec{x})$ and $P(\vec{x})$ are known, where the former is prescribed and the latter is computed with \texttt{firedrake} as stated above. The forward problem then becomes taking $a(\vec{x})$ as input into Equation~\eqref{darcy_pde} and computing the residual force balance, i.e.:
\begin{align}
\vec{r} = f(\vec{x}) + \nabla \cdot \left(a(\vec{x}) \nabla P(\vec{x})\right).
\end{align}
In other words, we aim to estimate the permeability field that minimizes the force balance residual for an observed pressure field, which is a strategy widely used in physics-informed machine learning \cite{raissi2019physics}. This approach removes the need for backpropagating through a numerical solver and facilitates comparison of our approach with methods that require gradients through the log posterior of $\theta(\vec{x})$. Each spatial field is resampled to a $100 \times 100$ square grid such that the gradients in Equation~\eqref{darcy_pde} can be evaluated with finite differences.

\subsubsection{Spline Representation}

\begin{figure}[ht]
\begin{center}
\includegraphics[width=0.85\textwidth]{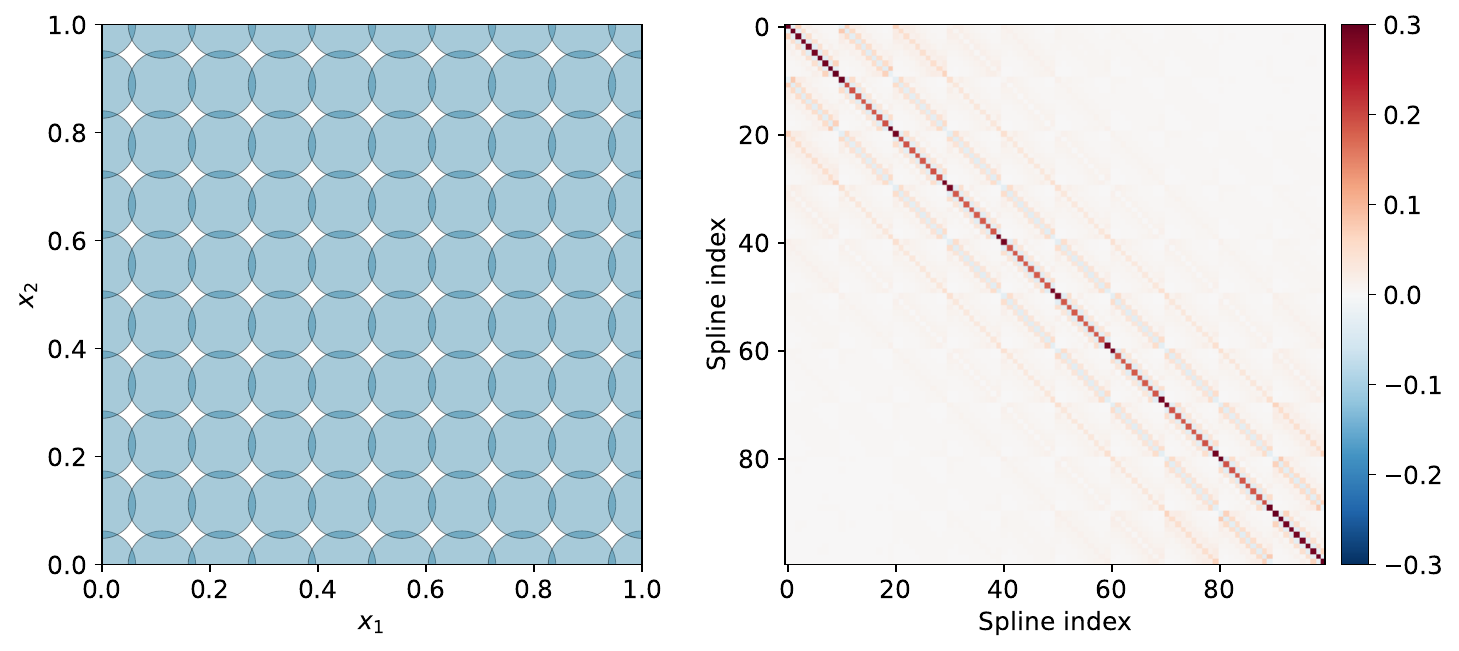}
\caption{\textbf{Left:} Illustration of the location and approximate supports for the 2D B-splines used for decomposing spatial fields. \textbf{Right:} Prior covariance matrix based on the Laplacian operator on the spline coefficients.}
\label{fig_splines_2d}
\end{center}
\end{figure}

We represent the $\theta(\vec{x})$ field as a linear combination of $100$ 2D B-splines distributed uniformly in both the horizontal and vertical directions (Figure \ref{fig_splines_2d}). 
The two-dimensional B-splines are simple outer products of the one-dimensional B-splines used in the linear regression problem in Section \ref{moderate_dim_problems} and can decompose spatial fields in a manner similar to wavelet transforms. Splines of multiple length scales can be combined in order to accurately represent multiscale fields. Here, we simply use a single length scale for our analysis.
Each spline is evaluated on the same $100 \times 100$ square grid as the pressure and source fields, flattened to a column vector, and placed into a column of a matrix $\vec{G} \in \mathbb{R}^{10000 \times 100}$. Overall, while the spline representation significantly reduces the dimensionality of the inverse problem, it still provides a bridge between the low- and moderate-dimensional problems investigated in the previous sections and higher-dimensional problems common in physical applications. Moreover, the spline representation admits a physically-motivated prior covariance matrix, which we now discuss.

The posterior distribution for the spline coefficient vector, $\vec{m}$, is (here we use $\vec{m}$ to avoid confusion with the coordinate space defined by $\vec{x}$):
\begin{align}
p(\vec{m} | \vec{r}) \sim p(\vec{r} | \vec{m}) p(\vec{m}).
\end{align}
In this experiment, we prescribe the likelihood $p(\vec{r} | \vec{m})$ as a Laplace distribution $\vec{r} = \mathcal{L}(0, b_r)$ where $b_r = 5.4$ is the scale parameter. The Laplace distribution permits a longer tail of $\vec{r}$ values while also providing a demonstration of Bayesian inference with non-Gaussian distributions. The prior distribution is represented as a multivariate Gaussian with a zero mean vector and a covariance matrix $\vec{\Sigma}_{\text{prior}}$. While the prior covariance can again be constructed with the squared-exponential Gaussian Process kernel to enforce a correlation length scale, we choose to instead construct a Laplacian smoothing matrix that estimates the 2nd-derivative of the spline coefficients in both directions (Figure \ref{fig_splines_2d}). This matrix format is well-suited to sparse representations, which can improve computational efficiency in higher dimensions, while still encouraging a spatially smooth $\theta(\vec{x})$. The spatial gradient of each spline in $\vec{G}$ is computed with finite differences in both directions, e.g. $\vec{G}_x$ and $\vec{G}_y$ are matrices containing the $x$- and $y$- gradients of the splines in their columns. The prior covariance is then computed as
\begin{align}
\Sigma_{\rm{prior}} = \lambda (\vec{G}_x^T \vec{G}_x +\vec{G}_y^T \vec{G}_y ) + \sigma_p^2 \vec{I},
\label{eqn_cov_prior}
\end{align}
where $\lambda$ is a scalar smoothing penalty and $\sigma_p^2 \vec{I}$ is an additive prior on the spline amplitudes. Here, we set $\lambda = 0.001$ and $\sigma_p = 1$.

\subsubsection{Sampling the Posterior Distribution}

Using the insights from the previous section, we apply our ensemble method with an Ornstein-Uhlenbeck forward process (Table \ref{table_esfs_high_dim}) to sample from the posterior distribution of the spline coefficients, where the scaling matrix of the diffusion process is defined by the Cholesky decomposition of the prior covariance matrix in Equation \ref{eqn_cov_prior}. As with the 20-dimensional problem in Section \ref{moderate_dim_problems}, posterior samples are also obtained with PMC, MALA, and NUTS for comparison. We omit EKS and ESS from this comparison since the former is restricted to Gaussian likelihoods and the latter requires a prohibitive number of parallel chains and forward evaluations for suitable performance.

\begin{figure*}[t!]
\begin{center}
\includegraphics[width=1.0\textwidth]{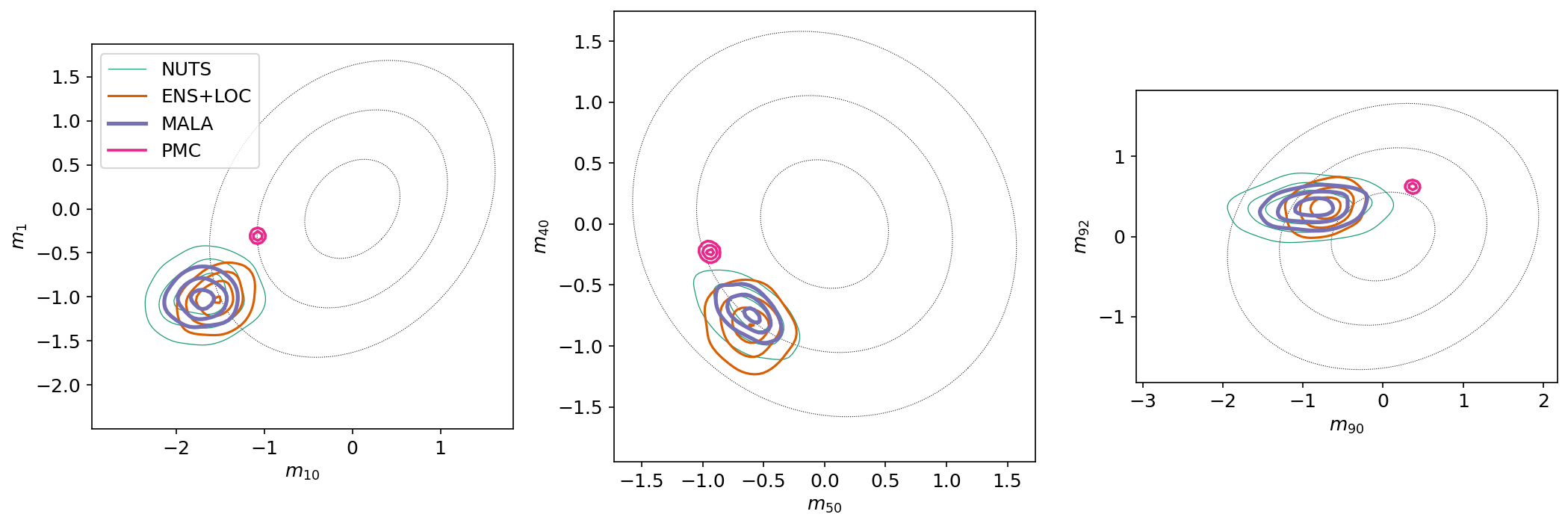}
\caption{\textbf{Marginal posterior kernel density estimates} for select pairs of two-dimensional B-splines used to reconstruct a permeability field given observations of a pressure field. Our ensemble method with an Ornstein-Uhlenbeck localization process (ENS+LOC) is compared to NUTS, MALA, and PMC. The black dashed ellipses in each plot correspond to the prior covariance (1-, 2-, and 3-$\sigma$). For this $100$-$d$ problem, the posterior samples obtained with ENS+LOC are similar to those obtained with the gradient-based NUTS and MALA.}
\label{fig_100d}
\end{center}
\end{figure*}

Compared with samples obtained from NUTS and MALA, our method successfully samples from the posterior distribution, although, similar to MALA, the samples exhibit slightly lower variance than those obtained with NUTS, see Figure \ref{fig_100d}. 
\begin{table}[ht]
\centering
\begin{tabular}{lccc}
\hline
Method & Energy distance & $N_{\text{fev}}$ & $N_{\text{gev}}$ \\
\hline
NUTS & - & $1 \times 10^6$ & $1 \times 10^6$ \\
MALA & 0.396 $\pm$ 0.039 & $4 \times 10^6$ & $4 \times 10^6$ \\
PMC & 12.269 $\pm$ 0.205 & $4 \times 10^6$ & -\\
\textbf{ENS+LOC} & \textbf{0.707 $\pm$ 0.050} & $6 \times 10^4$ & -\\
\hline
\end{tabular}
\caption{\textbf{Energy distance} and number of evaluations of the forward model, $N_{\text{fev}}$, and its gradient, $N_{\text{gev}}$, for different sampling methods for the permeability field experiment. Energy distance is computed relative to samples obtained with NUTS.}
\label{tab:energy_distance_100d}
\end{table}
This effect is likely driven by the relative weakness in the correlations (off-diagonal terms) of the prior covariance matrix used in the Ornstein-Uhlenbeck forward process, which in turn provides weaker constraints on the ensemble random walk during reverse diffusion. As a result, the likelihood ratio  $p_0(\vec{x}') / p_{\rm{is}}(\vec{x}')$ in Equation~\eqref{eqn_follmer_drift_mc} becomes dominated by a small subset of the ensemble members, causing the other members to drift towards that subset. We find this effect to be more pronounced when using scalar (isotropic) diffusion processes. Note that this effect also drives the relatively poor performance of methods like PMC, where samples collapse to a point that is generally offset from the posterior mean. Overall, our ensemble method with the Ornstein-Uhlenbeck forward process is successfully able to sample a substantial region of the posterior distribution in high-dimensions without requiring gradients of the distribution. Moreover, it is able to obtain competitive sample accuracy with significantly fewer evaluations of the forward model (Table \ref{tab:energy_distance_100d}).

\subsection{Implementation Details, Performance, and Computational Considerations}
\label{section_computation}

The practical realization of the proposed gradient-free sampling method involves simulating the reverse diffusion SDE (Equation \ref{eqn_rev_process}) while approximating the required score function $s_t(x)$ using an evolving particle ensemble $\left\{x_i\right\}$. The estimation employs importance sampling where the proposal distribution $p_\text{is}$ is adaptively derived from the ensemble's current statistics (e.g., mean and covariance, Equation \ref{eqn_esfs_is}), potentially enhanced by variance reduction techniques like antithetic sampling (Equation \ref{eqn:antithetic}). In this way, ensemble member interaction is collective during the estimation of $s_t(x)$, which then governs the drift for all ensemble members in the reverse diffusion SDE. To manage computational expense, particularly when target density evaluations are costly, the importance sampling distribution $p_\text{is}$ and corresponding score estimate $\hat{s}_t$ are often updated only periodically (at $N_r$ intervals) rather than at every SDE time step (Section \ref{section_periodic_resampling}). The number of score updates, $N_r$, presents a trade-off between computational budget (total target evaluations scale as $N_\text{ens} * N_r$) and the fidelity of the score approximation over time. Fewer updates reduce cost but increase reliance on a potentially inaccurate score estimate. Naturally, the SDE discretization step $dt$ must also be chosen appropriately relative to the score update frequency ($\Delta t_r = T / N_r$) to ensure numerical stability.

The accuracy and efficiency of this approach are intrinsically linked to several factors. The ensemble size, $N_\text{ens}$, is the most important factor, as the variance of the Monte Carlo score estimate scales asymptotically as $\mathcal{O}(1/ N_\text{ens}$) (\ref{appendix_convergence_analysis}). This establishes $N_\text{ens}$ as the primary control for the accuracy of the score approximation and the resulting SDE dynamics. However, sufficient ensemble sizes are themselves dictated by the dimensionality of the problem. For low- to moderate-dimensional distributions, and similar to the ensemble Kalman methods \citep[][]{garbuno2020interacting}, $N_\text{ens}$ should at least be 10--100 times the dimensionality. One can empirically verify the sufficiency of the ensemble size by monitoring the stability of the final sample statistics (mean, covariance) across runs with increasing $N_\text{ens}$. Additionally, visualizing the evolution of marginal distributions of the ensemble during the reverse diffusion path can provide qualitative insights into whether the ensemble is moving towards the target or exhibiting signs of collapse or excessive variance due to insufficient $N_\text{ens}$.

At higher dimensions (e.g., higher than 15--20), the isotropic diffusion term will likely lead to low importance weights for most ensemble members. Performance improvement can then be realized by employing anisotropic Ornstein-Uhlenbeck diffusion (Section \ref{subsection_ou_fwd}). The OU process leverages prior information ($\mu$, $\Sigma_\text{prior}$, scaled by $\alpha \ge 1$) to guide exploration and constrain the diffusion, improving the relevance of the ensemble for score estimation. This approach exploits structure in the target distribution via the prior, which is roughly consistent with improvements for NUTS and MALA that exploit local structure in the posterior to guide proposal steps for their respective sampling dynamics \citep[][]{girolami2011riemann}. The OU reversion parameter $\theta$ modulates the influence of the prior, with smaller values (e.g., 0.1--0.2) accommodating larger discrepancies between prior and posterior. However, uninformative priors will cause the OU process to behave like the base isotropic process, likely leading to lower importance sampling weights for higher dimensions. Similarly, posterior distributions that are substantially different from the prior (e.g., highly non-Gaussian) will also likely lead to lower importance weights for higher dimensions.

The interplay between initial ensemble diversity and the non-linearity of the target distribution is an important consideration (Section \ref{section_ensemble_init}). For relatively smooth, near-Gaussian targets, our method demonstrates robustness even if the initial ensemble has only moderate overlap with the true $p_0$; the adaptive importance sampling can still yield effective score estimates to guide the ensemble towards the high-probability regions. However, for highly non-linear, skewed, or multi-modal distributions, the quality of the score estimation becomes more sensitive to the initial ensemble placement and diversity (Figure \ref{fig_rev_banana}). If the initial overlap is poor, the importance weights $p_0(x') / p_\text{is}(x')$ can exhibit high variance across the ensemble members, potentially leading to less accurate score estimates and slower convergence or even undersampling of certain modes. This effect highlights that while the method is designed to handle non-Gaussian behavior (as demonstrated in our experiments), ensuring sufficient initial diversity, particularly for complex target landscapes, contributes to more reliable and efficient sampling.

Consequently, our method demonstrates optimal performance when analytic gradients are intractable but target evaluations are feasible (and potentially parallelizable), and particularly in low-to-moderate dimensions or when informative priors enable effective anisotropic diffusion. The primary limitation remains the curse of dimensionality impacting the Monte Carlo score estimation, requiring larger $N_\text{ens}$ and potentially struggling with highly complex target geometries poorly approximated by the chosen importance sampling scheme.

From a computational cost perspective, the dominant expense is typically the $N_\text{fev} = N_\text{ens} * N_r$ target density evaluations. This contrasts with gradient-based MCMC (NUTS, MALA) which require both log probability and gradient calls, where the number of steps can vary greatly depending on target complexity and mixing properties. While direct comparison is nuanced, our approach often necessitates similar or fewer forward model calls than NUTS/MALA in our various experiments, while avoiding gradient computations. The cost scales similarly to Ensemble Kalman methods (ensemble size * number of iterations). However, the need for increased $N_\text{ens}$ in higher dimensions can elevate the cost relative to gradient-based techniques, unless the cost of gradient computation is substantially higher than that of the target evaluation itself. Thus, the proposed method offers a competitive advantage primarily when gradient computation is the main bottleneck or when its ensemble nature provides benefits in exploring specific target structures like multimodality.

\section{Conclusions}
\label{conclusions}
In this study, we introduced the ensemble strategies that leverage the dynamics of particle ensembles to estimate a score function used to generate samples from a target distribution within a reverse diffusion process. This method, rooted in principles of importance sampling, is robust for low- to medium-dimensional problems where gradients of the posterior log probability or the forward model are unavailable. This situation is often the case for complex black box forward models where analytic gradients are intractable and automatic differentiation methods are not an option. Furthermore, many real-world systems exhibit inherent non-differentiabilities, such as abrupt phase transitions, stochastic components, or discrete events, which render gradient-based methods inapplicable.

Our experiments, encompassing a hierarchy of problems with a range of dimensionalities, highlight the advantages of our method over traditional MCMC methods such as NUTS and MALA, even when these methods use gradient information. Particularly noteworthy is the ability to effectively navigate and sample from non-Gaussian distributions, which makes this approach an appealing choice especially in fields like geophysics and computational physics where such challenges are prevalent.

Furthermore, the implementation of resampling strategies has substantially enhanced the computational efficiency of this method when compared to related gradient-free strategies such as vanilla Föllmer sampling. The adaptability of our method to various problem sizes and its ability to integrate domain-specific knowledge (e.g., using prior covariances in geophysical applications) further underscore its practical utility.

The ideas presented here are however not without challenges and warrant further investigation. As demonstrated across our experiments, the efficiency of our method and the required ensemble size are dependent on the dimensionality of the target distribution, a common characteristic of Monte Carlo-based approaches. The scalability to very high-dimensional problems remains a challenge that is rooted in the obvious curse of dimensionality. While our strategy of utilizing Ornstein-Uhlenbeck processes defined by prior information has shown promise, additional research is needed to optimize for such scenarios. In other Bayesian inference scenarios, priors may be relatively uninformative, which could limit the utility of the Ornstein-Uhlenbeck processes. In such cases, it would be advantageous to discover local structure in the posterior rather than relying on global structure from the prior. Other works have approached this problem by training neural network models of the score, but those methods require taking gradients of the log posterior. It remains a challenge to build a completely gradient-free method that also works in high dimensions while remaining computationally efficient.

\section*{CRediT authorship contribution statement}
\textbf{Bryan Riel:} Writing - review \& editing, Writing - original draft, Visualization, Validation, Software, Formal analysis. \textbf{Tobias Bischoff:} Writing - review \& editing, Writing - original draft, Conceptualization, Methodology, Software

\section*{Software and Data}
We implemented all algorithms and strategies presented in this work in JAX \cite{jax2018github}. In particular, we used the Equinox \cite{kidger2021equinox} and Diffrax \cite{kidger2022neural} packages to build a flexible implementation of forward-reverse diffusion systems. 

\section*{Acknowledgments}
We thank the Editor in Chief, Associate Editor, and two anonymous reviewers for their constructive comments for improving this work. This work was supported by projects funded by the National Natural Science Foundation of China (42376230) and the Zhejiang University Global Partnership Fund.

\appendix
\section{Energy distance}
\label{appendix_energy}
The energy distance $\epsilon$ is related to the covariance distance between two random vectors $\vec{X}$ and $\vec{Y}$ and is defined as \cite{szekely2013energy}:
\begin{align}
\varepsilon(\vec{X}, \vec{Y}) = 2\mathbb{E}(\|\vec{X} - \vec{Y}\|_p) - \mathbb{E}(\|\vec{X} - \vec{X}'\|_p) - \mathbb{E}(\|\vec{Y} - \vec{Y}'\|_p),
\end{align}
where $\|\cdot\|_p$ is the $p$-norm and $\vec{X}'$ and $\vec{Y}'$ are random permutations of $\vec{X}$ and $\vec{Y}$, respectively. As such, the estimate for $\varepsilon(\vec{X}, \vec{Y})$ can be repeated for multiple realizations of $\vec{X}'$ and $\vec{Y}'$ such that the mean value represents a more robust estimate. Here, we use the Python package \texttt{dcor} package to compute $\varepsilon(\vec{X}, \vec{Y})$ \cite{ramos-carreno+torrecilla_2023_dcor}. As this estimate can vary depending on the size of the vectors $\vec{X}$ and $\vec{Y}$, comparing the samples generated by our method for different ensemble sizes requires multiple runs of our method with different initial random ensembles until a vector of a fixed size $N_{\text{total}}$ is filled, where $N_{\text{total}}$ is several multiples of $N_{ens}$. Additionally, energy distance can be computed for different random permutations, allowing us to assess the statistics of sample accuracy. Overall, we find the energy distance to be a robust way of comparing samples from arbitrary probability distributions, and we find that using a $p$-norm with $p = 1$ or 1.5 can mitigate the effects of outliers in $\vec{X}$ and $\vec{Y}$.

\section{Multiple Importance Sampling Approach}
\label{appendix_mis}
\begin{figure}[t]
\begin{center}
\includegraphics[width=0.5\textwidth]{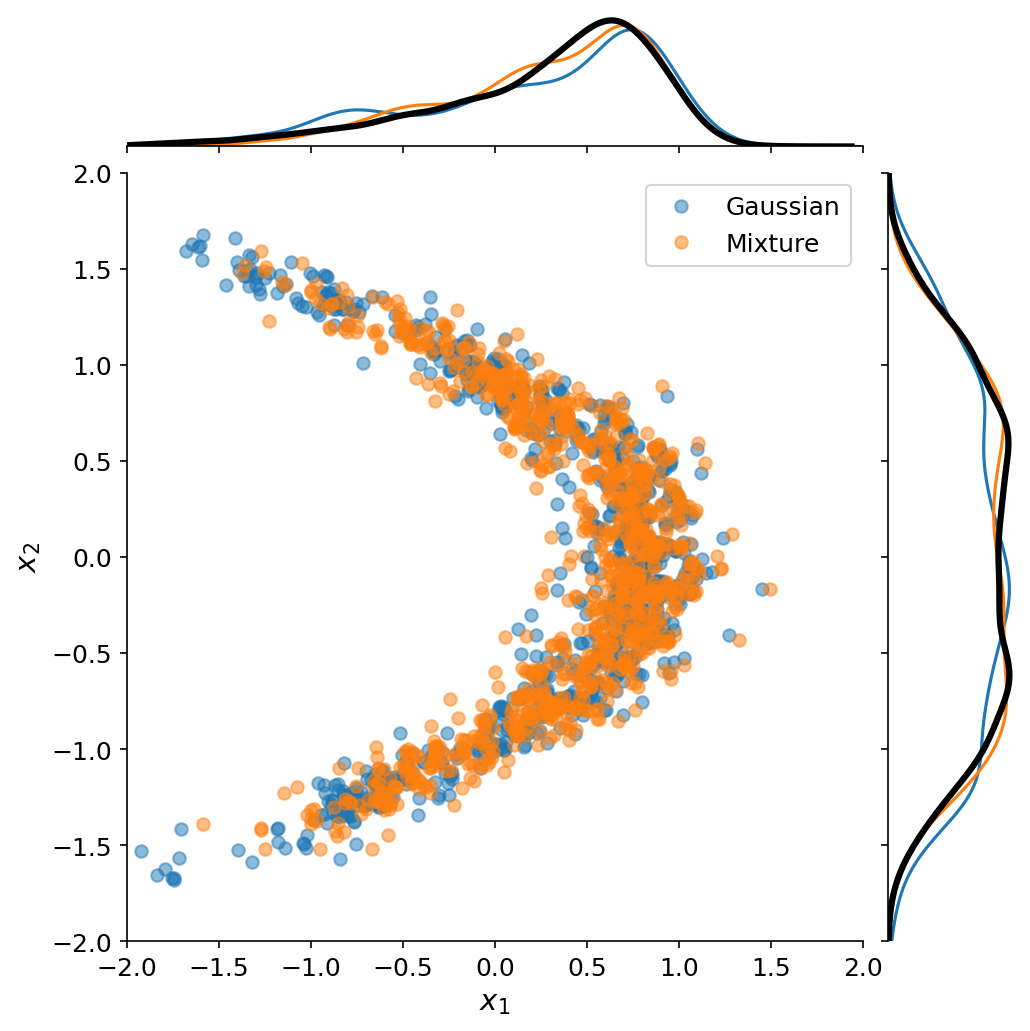}
\caption{Comparison of samples from the ``banana'' distribution using either the \textbf{Gaussian importance sampling} distribution (blue) or the \textbf{multiple importance sampling} distribution corresponding to a mixture of Gaussians (orange). The black line in the marginal plots indicates the ground truth probability density. While both importance sampling distributions result in accurate samples from the ``banana'' distribution, the mixture distribution leads to probability densities that are slightly closer to the ground truth.}
\label{fig_banana}
\end{center}
\end{figure}

An alternative to the simple Gaussian importance sampling distribution $p_{\rm{is}}$ from Section~\ref{sec_imp_sampling} is to use a multiple importance sampling strategy (e.g., \cite{elvira2022advances}), where each component is a Gaussian centered on the location of an ensemble member. The multiple importance sampling estimator of $p_t$ is then given by
\begin{equation}
        \hat{p}_t(\vec{x}) = \sum_{i=1}^{N_{\rm{ens}}} \frac{1}{N_i} \sum_{j=1}^{N_i} \kappa_t(\vec{x}|\vec{x}'_{ij}) w_i(\vec{x}'_{ij})\frac{p_0(\vec{x}'_{ij})}{p_{{\rm{is}}, i}(\vec{x}'_{ij})}, \,\, \vec{x}'_{ij} \sim p_{{\rm{is}}, i},
\end{equation}
where $N_i$ is the number of samples drawn for component $i$. By choosing $p_{{\rm{is}}, i}(\vec{x}) = \kappa_t(\vec{x}|\vec{x}_i)$, where $\vec{x}_i$ is an ensemble member (e.g., a sample in the process of being generated by the reverse diffusion process from Equation~\eqref{eqn_rev_process}), and by computing $w_i$ using the balance heuristic \cite{elvira2022advances}
\begin{equation}
    w_i(\vec{x}) = \frac{N_i\kappa_t(\vec{x}|\vec{x}_i)}{\sum_{j=1}^{N_{\rm{ens}}} N_j \kappa_t(\vec{x}|\vec{x}_j)}, 
\end{equation}
one can assemble a multiple importance sampling estimator based on many Gaussians with time-varying properties. At $t=1$, the Gaussians are wide and overlapping and near $t=0$ they approach delta peaks. If the number of samples drawn for each importance sampling distribution is $N_i = 1, \forall i$, then the estimator $\hat{p}_t$, simplifies to
\begin{equation}
    \hat{p}_t(\vec{x}) = \sum_{i=1}^{N_{\rm{ens}}} \kappa_t(\vec{x}|\vec{x}'_i) \frac{p_0(\vec{x}'_i)}{\sum_{j=1}^{N_{\rm{ens}}}\kappa_t(\vec{x}'_i|\vec{x}_j)}, \,\, \vec{x}'_i \sim \kappa_t(.|\vec{x}_i), 
\end{equation}
which resembles an importance sampling estimator where the importance sampling distribution is a mixture of Gaussians with equal weights.
We found that this estimator can be favorable over the simple Gaussian one presented in the main text for problems that are highly non-Gaussian, e.g., the ``banana'' distribution in Figure~\ref{fig_banana}.

\section{Convergence Analysis}
\label{appendix_convergence_analysis}
We begin by writing the approximate solution to the forward noising process as
\begin{equation}
    \hat{p}_t(\vec{x}) = p_t(\vec{x}) + \vec{\epsilon}_t(\vec{x}), 
\end{equation}
where the Monte Carlo error is unbiased, $\mathbb{E}\left[\vec{\epsilon}_t\right] = \vec{0}$, and its variance decreases with the number of samples $N$ as $\mathbb{E}\left[\vec{\epsilon}_t^2\right] = \frac{C_d}{N}$ (using the notation $N_{\text{ens}} = N$ for brevity). Here, $C_d$ is a constant that depends on the dimensionality of the problem and on the importance sampling distribution used to compute the Monte Carlo estimate. The approximate score function can be partitioned into an exact score function and an error term
\begin{equation}
    \hat{\vec{s}}_t(\vec{x}) = \vec{s}_t(\vec{x}) + \vec{\epsilon}_t(\vec{x}), 
\end{equation}
where we re-use the notation for the Monte Carlo error for $\hat{p}_t(\vec{x})$. As $N$ increases, the behavior of this error follows the same as for $\hat{p}_t(\vec{x})$, which can be seen by a Taylor expansion of $\hat{p}_t(\vec{x})$ inside the expression for $\hat{\vec{s}}_t(\vec{x})$. The error introduced by the Monte Carlo sampling procedure for the score function is hence unbiased and its variance decreases with $N^{-1}$ as $N \rightarrow \infty$.

If we base the error function for the sampling procedure on the denoising score matching loss \cite{song2021scorebased, song2019generative}, we can write
\begin{subequations}
    \begin{align}
        \mathcal{L}_N(\vec{\epsilon)} &= \mathbb{E}_{t\sim U[0,1]}\left[ \mathbb{E}_{\vec{x}_0\sim p_0}\left[\mathbb{E}_{\vec{z}\sim \mathcal{N}(\vec{0}, \vec{1})}\left[ \left(\Sigma_t\hat{\vec{s}}_t(\vec{x}_0 + \Sigma_t\vec{z}) + \vec{z} \right)^2\right]  \right] \right]\\
        &= \mathbb{E}_{t\sim U[0,1]}\left[ \mathbb{E}_{\vec{x}_0\sim p_0}\left[\mathbb{E}_{\vec{z}\sim \mathcal{N}(\vec{0}, \vec{1})}\left[ \left(\Sigma_t\vec{s}_t(\vec{x}_0 + \Sigma_t\vec{z}) + \vec{z} \right)^2\right]  \right] \right]\\
        &+ 2\mathbb{E}_{t\sim U[0,1]}\left[ \mathbb{E}_{\vec{x}_0\sim p_0}\left[\mathbb{E}_{\vec{z}\sim \mathcal{N}(\vec{0}, \vec{1})}\left[ \left(\Sigma_t\vec{\epsilon}_t(\vec{x}_0 + \Sigma_t\vec{z})\right)^T\vec{z}\right]  \right] \right]\\ 
        &+ \mathbb{E}_{t\sim U[0,1]}\left[ \mathbb{E}_{\vec{x}_0\sim p_0}\left[\mathbb{E}_{\vec{z}\sim \mathcal{N}(\vec{0}, \vec{1})}\left[ \left(\Sigma_t \vec{\epsilon}_t(\vec{x}_0 + \Sigma_t\vec{z})\right)^2\right]  \right] \right].
    \end{align}
\end{subequations}
We can then take the expectation over random realizations of the Monte Carlo sampling error $\vec{\epsilon}$ and use the fact that the error is component-wise unbiased with component-wise variance that asymptotically behaves as $N^{-1}$ as $N \rightarrow \infty$. We thus come to the conclusion that
\begin{equation}
    \mathcal{L}_N = \mathbb{E}_{\vec{\epsilon}}\left[ \mathcal{L}_N(\vec{\epsilon)} \right] \sim \frac{1}{N}, \,\, N \rightarrow \infty.
\end{equation}
This result confirms that the average denoising score matching error tends to zero as the number of ensemble members $N$ increases to infinity (while reiterating that the prefactor in the asymptotic relationship depends on the dimensionality of the problem and on the importance sampling distribution). Consequently, based on the $~1/N$ convergence for the average score matching loss, the accumulated error propagated through the reverse diffusion process implies that the deviation of the final generated sample distribution $\hat{p}_0$ from the target $p_0$, measured under an appropriate metric, is expected to decrease asymptotically as $~1/\sqrt{N}$ \citep[][]{robert1999monte}. For a finite $N_\text{ens}$ and finite number of integration steps, some bias in the final samples relative to $p_0$ may be present due to the cumulative effect of using an approximate score (see Section \ref{section_computation} for further discussion). While this bias is systematically reduced by increasing $N_\text{ens}$, formalizing the bias for an arbitrary distribution remains difficult and currently relies on empirical validation (e.g., Figure \ref{fig_ensemble_size}).

\section{Additional Gradient-free Sampling Algorithms}
\label{appendix_gradient_free_algo}

In addition to sampling methods such as Ensemble Kalman Sampling and our ensemble-based diffusion method, other algorithms have been developed to sample from probability distributions without requiring gradients of the distribution. Many of these algorithms also involve the evolution of sample ensembles through dynamics governed by the log probabilities of the samples. Here, we explore two additional techniques that exhibit complementary characteristics to EKS and our method. The first is Ensemble Slice Sampling \citep[][]{karamanis2021ensemble} (ESS), which updates ensemble members in parallel by defining intervals or hyperrectangles based on the current slice level, where the boundaries for sampling new points are adaptively constructed using locations of other ensemble members. This method has been shown to be performant for distributions with a low to moderate number of dimensions. The second is Population Monte Carlo \citep[][]{cappe2004population} (PMC), which belongs to the family of adaptive importance sampling techniques. PMC works by adapting an importance sampling distribution to the statistics of the ensemble, which is similar to the resampling strategy employed by our method (Section \ref{section_periodic_resampling}). One key difference is that PMC adapts to the \textit{fixed} target distribution, whereas our strategy adapts to a \textit{time-dependent} distribution as governed by the reverse diffusion.

We use ESS as implemented in NumPyro \citep[][]{pyro,numpyro} and use twice the number of chains as the dimensionality of the distribution. For PMC, we adapt a Gaussian mixture proposal distribution for a given number of iterations. The proposal distribution mean, $\mu_i$, and covariance, $\Sigma_i$, for mixture component $i$ are adapted with a damping rate, $\gamma$, such that $\mu_i \leftarrow (1-\gamma)\mu_i+\gamma\,\hat{\mu}_i$ and $\Sigma_i \leftarrow (1-\gamma)\Sigma_i+\gamma\,(\hat{\Sigma}_i+\epsilon I)$. Here, $\hat{\mu}_i$ and $\hat{\Sigma}_i$ are the ensemble-estimated mean and covariance for the $i$-th component, and $\epsilon I$ is a diagonal jitter matrix for mitigating ensemble collapse \citep[][]{cappe2004population}. Generally, a lower value of $\gamma \in [0, 1]$ coupled with the jitter term is necessary for sampling from distributions with higher dimensions by allowing the mixture proposal to adapt to the ensemble more gradually.

\begin{figure}[ht]
\begin{center}
\includegraphics[width=1.0\textwidth]{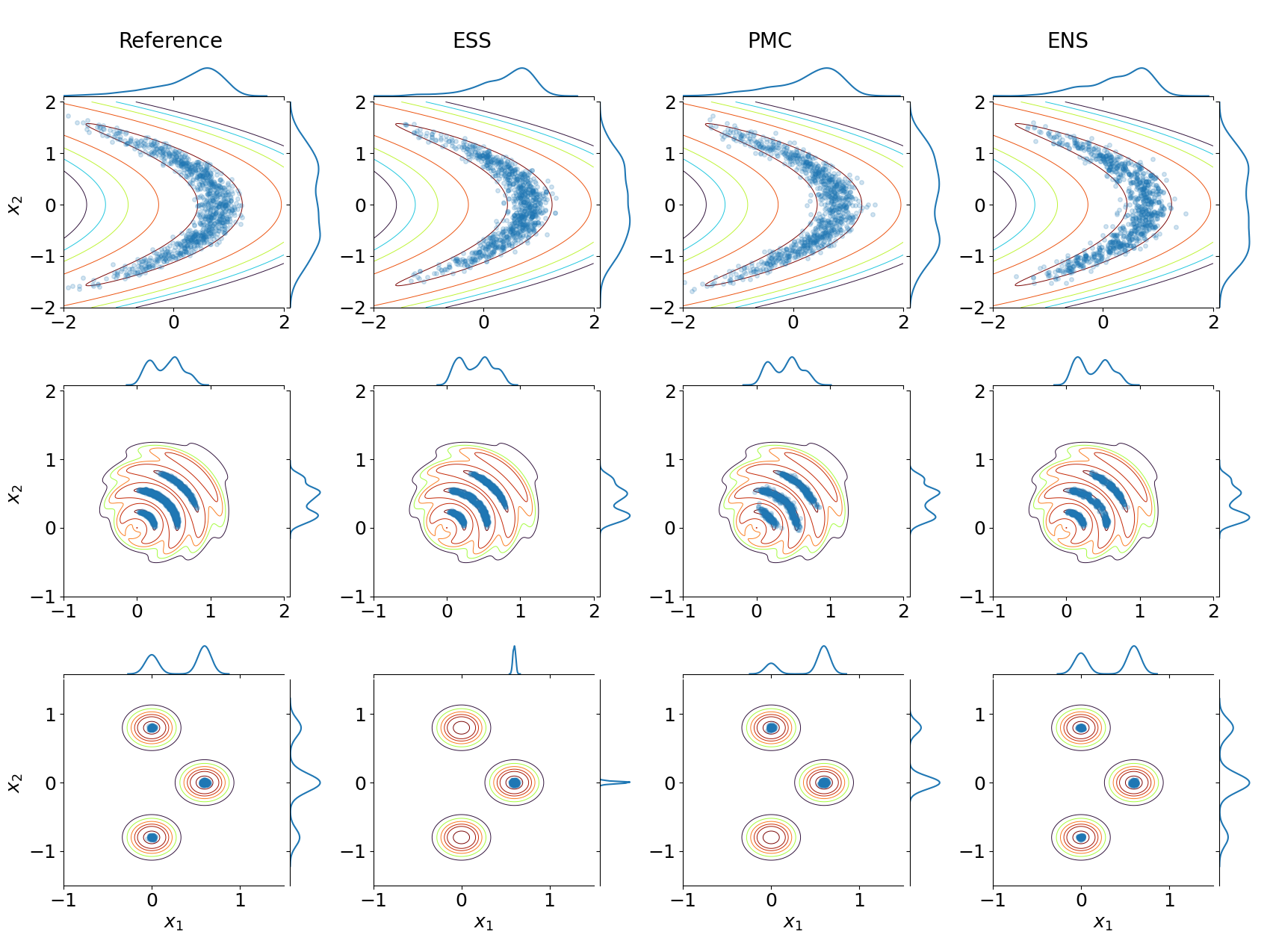}
\caption{Similar comparison as Figure \ref{fig_2d_comparison} but with two additional gradient-free sampling techniques: Ensemble Slice Sampling (ESS) and Population Monte Carlo (PMC). All three gradient-free sampling methods perform well for the banana and ridged probability distributions. However, ESS tends to only sample from a single mode for the mixture distribution, while PMC tends to sample from either one or two modes.}
\label{fig:appendix_2d_gradientfree}
\end{center}
\end{figure}

Figure \ref{fig:appendix_2d_gradientfree} show the samples generated by ESS and PMC for the two-dimensional problems in Figure \ref{fig_2d_comparison}, compared to those generated by our ensemble methods, ENS. ESS and PMC perform well for the banana and ridge distributions since both are not restricted by any assumptions on the form of the target distribution. However, both still have difficulties sampling all three modes for the mixture distribution.

%%       \citet{<label>} ==> Lamport [21]
%%       \citep{<label>} ==> [21]
%%

\bibliographystyle{elsarticle-num-names} 
\bibliography{ref-mod}


\end{document}